\documentclass[onecolumn,12pt]{article}
\usepackage{enumitem}
\usepackage{subcaption}
\usepackage[top=1.2in, bottom=1.2in, left=1in, right=1in]{geometry}
\usepackage{amsmath,amsfonts,amscd,amssymb}
\usepackage{graphicx}
\usepackage{mathdots}
\usepackage{epstopdf}
\usepackage{psfrag}
\usepackage{overpic}
\usepackage{cancel}
\usepackage{rotating}
\usepackage{url}
\usepackage{caption}
\usepackage{color}
\usepackage{rotating}
\usepackage{multirow}
\usepackage{wrapfig}
\usepackage{mathtools}
\usepackage{subeqnarray}
\usepackage{setspace}
\usepackage{palatino} %
\usepackage[numbers,sort&compress]{natbib}
\usepackage[table]{xcolor}    %
\usepackage[numbered,framed]{matlab-prettifier}
\usepackage{booktabs}
\usepackage{authblk}

\usepackage{array}
\newcolumntype{L}{>{\centering\arraybackslash}m{10cm}}

\usepackage[bottom,flushmargin,hang,multiple]{footmisc}
\usepackage{lipsum}
\newcommand\blfootnote[1]{%
  \begingroup
  \renewcommand\thefootnote{}\footnote{#1}%
  \addtocounter{footnote}{-1}%
  \endgroup
}

\definecolor{header1}{cmyk}{0,0,0,1}

\usepackage{amsthm}

\usepackage{mleftright}
\usepackage{amssymb}
\usepackage{lipsum}
\usepackage{tikz,pgfplots,pstool}
\usetikzlibrary{shapes.misc}
\tikzset{cross/.style={cross out, draw=blue, minimum size=2*(#1-\pgflinewidth), inner sep=0pt, outer sep=0pt},
	cross/.default={1pt}}

\pgfplotsset{compat=1.13}

\usepackage{contour}
\usepackage[normalem]{ulem}

\contourlength{0.8pt}

\newcommand{\myuline}[1]{%
  \uline{\phantom{#1}}%
  \llap{\contour{white}{#1}}%
}

\usepackage{subcaption}

\newtheoremstyle{mylemmastyle}
  {3pt} %
  {3pt} %
  {\itshape} %
  {} %
  {\bfseries} %
  {.} %
  {\newline} %
  {\thmname{#1} \thmnumber{#2}: \thmnote{\normalfont\myuline{#3}}} %

\theoremstyle{mylemmastyle}

\usepackage{algorithm}
\usepackage{algorithmicx}
\usepackage{algpseudocode}
\algrenewcommand{\algorithmiccomment}[1]{\hfill #1}

\usepackage{hyperref}
\usepackage{xcolor}
\hypersetup{
    colorlinks,
    linkcolor={red!70!black},
    citecolor={blue!70!black},
    urlcolor={green!50!black}
}
\definecolor{col0}{RGB}{82,173,2}
\definecolor{col1}{RGB}{212,173,2}
\definecolor{col2}{RGB}{255, 111, 0}

\title{\vspace{-.4in}\textbf{Calibrating Adaptive Smoothing Methods for Freeway Traffic Reconstruction}\vspace{-.1in}}
\author[1,2]{Junyi Ji$^*$}
\author[2]{Derek Gloudemans}
\author[2]{Gergely Zach\'ar}
\author[3]{Matthew W. Nice}
\author[2]{William Barbour}
\author[1,2]{Daniel B. Work}

\affil[1]{\small Department of Civil and Environmental Engineering, Vanderbilt University}
\affil[2]{\small Institute for Software Integrated Systems, Vanderbilt University}
\affil[3]{\small Virginia Tech Transportation Institute (VTTI), Virginia Polytechnic Institute and State University}

\date{}

\renewcommand{\vec}[1]{\boldsymbol{\mathrm{#1}}}

\renewcommand{\vec}[1]{\boldsymbol{\mathrm{#1}}}

\usepackage[font=footnotesize,labelfont=bf]{caption}
\begin{document}
\maketitle
\blfootnote{$^*$ Correspondence at \href{mailto:junyi.ji@vanderbilt.edu}{junyi.ji@vanderbilt.edu}}
\date{}
\vspace{-1.25cm}

\begin{abstract}
    The adaptive smoothing method (ASM) is a widely used approach for traffic state reconstruction. This article presents a Python implementation of ASM, featuring end-to-end calibration using real-world ground truth data. The calibration is formulated as a parameterized kernel optimization problem. The model is calibrated using data from a full-state observation testbed, with input from a sparse radar sensor network. The implementation is developed in PyTorch, enabling integration with various deep learning methods. We evaluate the results in terms of speed distribution, spatio-temporal error distribution, and spatial error to provide benchmark metrics for the traffic reconstruction problem. We further demonstrate the usability of the calibrated method across multiple freeways. Finally, we discuss the challenges of reproducibility in general traffic model calibration and the limitations of ASM. This article is reproducible and can serve as a benchmark for various freeway operation tasks.
\end{abstract}


\begin{spacing}{.9}
\small{
\setlength{\bibsep}{3.5pt}
\section{Introduction}
\label{sec1}

The adaptive smoothing method (ASM) \cite{treiber2011reconstructing} is a celebrated method for reconstructing spatio-temporal traffic patterns from sparse sensor data, especially for traffic waves \cite{helbing1998jams}. ASM was first introduced by Treiber and Helbing in 2002 \cite{treiber2002reconstructing, treiber2003adaptive} and later refined by Treiber and Kesting \cite{kesting2010datengestutzte, treiber2011reconstructing}. Earlier work by Treiber et al. in 2000 \cite{treiber2000congested} applied conventional kernel smoothing and interpolation, and the development of ASM afterwards was driven by insights from traffic wave phenomena and advances in filter design. Later in 2010, Schreiter et al. \cite{schreiter2010two,schreiter2013vehicle} proposed two fast implementations of ASM using Fast Fourier Transform (FFT) and cross-correlation, significantly improving computational efficiency.  

It can be used for interpolating and smoothing traffic data, and serves as an important benchmark~\cite{mohammadian2021performance} for various freeway operation tasks, including traffic speed estimation \cite{he2024efficient,wu2024traffic}, data refinement \cite{he2023refining,ji2025stop}, data imputation \cite{benkraouda2020traffic,yang2024advanced}, data preprocessing \cite{coursey2024ft,li2021multistep}, vehicle trajectory reconstruction \cite{chen2024macro}, and emission estimation \cite{wang2011estimating,tsanakas2022generating}. The method originated in Europe \cite{treiber2002reconstructing,treiber2003adaptive}, where it was widely used \cite{van2010robust, wang2011estimating,li2021multistep}, and has since been gradually adopted internationally. As summarized by \cite{schreiter2013vehicle}, ASM is suitable for dynamic traffic management applications. In the age of artificial intelligence (AI) applications, ASM can still play an important role in freeway operations, for example, data upsampling \cite{ji2025stop}, and even shows promise as a baseline for predictive modeling. A detailed list of applications of ASM can be found in Table~\ref{tab:literature}. 

\begin{table}[ht]
\centering
\footnotesize
\caption{Overview of ASM applications. Each row highlights key aspects of ASM usage, including task type, the presence of calibration, spatial scale, number of implementation locations, geographic location, and its code availability for reproducible research. NGSIM refers to the Next Generation Simulation dataset \cite{ngsim}, and ZTD refers to the Zen Traffic Data \cite{seo2020evaluation}, corridor-scale refers to the spatial coverage of over 10 kilometers.}
\label{tab:literature}
\begin{tabular}{cccccccc}
\hline
& Task & Calibration & Scale & Locations & Geographic & Code \\ \hline
\cite{van2010robust}   & Reconstruction       &      Yes     & Corridor-scale &Single &Netherlands&   No     \\
\cite{wang2011estimating} & Estimation    &    No         &   Corridor-scale   &Single & Netherlands &     No  \\ 
\cite{chen2018adaptive} & Estimation      &    Yes         &   Corridor-scale   &Single & China &     No  \\ 
\cite{benkraouda2020traffic} & Imputation        &     No         & NGSIM     &Single &U.S.&  No     \\ 
\cite{tsanakas2022generating} & Reconstruction     &     No         & NGSIM   &Single &U.S.&  No     \\ 
\cite{he2024efficient}   & Estimation   &      No       & NGSIM      &Single &U.S.&   No     \\
\cite{he2023refining}  &Refinement         &       No      &  ZTD    &Single &Japan &     No  \\
\cite{chen2024macro}  &Reconstruction    &       No      &  NGSIM   &Single &U.S. &     No  \\
\cite{yang2024advanced}& Reconstruction        &    Yes         &   NGSIM   &Single &U.S. &     No  \\ 
\cite{wu2024traffic} & Estimation     &    No         &   NGSIM   &Single & U.S. &     Yes  \\ 
\hline
Ours & -  &   Yes &Corridor-scale &Multiple & U.S.&    Yes \\ 
\hline
\end{tabular}
\end{table} 

The advantage of ASM lies in its simplicity on just six parameters and its grounding in empirical observations. It also offers flexibility and shows strong potential for widespread adoption \cite{chen2018adaptive} with appropriate calibration.  Despite these strengths, the method’s impact has been limited by the absence of publicly available, implementation-ready code for large-scale freeway networks. Table~\ref{tab:literature} summarizes the literature using ASM across various tasks, including both further development and as baseline for comparison. Despite its continued relevance and active use within the research community, a publicly available implementation of ASM is lacking. This absence potentially results in redundant efforts as research groups independently re-implement the method. Apart from our work, the only publicly available implementation we identified is by \cite{wu2024traffic}, who released their code via GitHub\footnote{The code is available at \url{https://github.com/Lucky-Fan/GP_TSE}.}. However, this implementation is limited to the Next Generation Simulation (NGSIM) dataset and is not yet suitable for applications at a larger scale.
This gap hinders reproducibility and slows uptake by the broader research community\footnote{The authors first attempted to implement the methods in November 2023; however, the initial implementation faced issues. After iterating through three different versions, we ultimately achieved a computational speedup for large-scale systems.}. Another challenge is the previous lack of full-state data necessary for the calibration. In practice, most studies adopt the parameter set suggested by Treiber and Kesting \cite{treiber2011reconstructing}. This article addresses the aforementioned challenges by calibrating the model using data collected from the I-24 MOTION testbed and providing a fast, trainable, and accurate implementation of ASM.

The contributions of this paper are summarized as follows:
\begin{enumerate}[label=(\roman*),noitemsep]
\item We provide a fast and calibrated implementation of the Adaptive Smoothing Method (ASM), supporting the end-to-end calibration and field implementation.
\item We calibrate the model using real-world radar-sensor data and ground-truth observations from the fully instrumented I-24 MOTION testbed, and comprehensively evaluate its performance. The resulting implementation serves as a reproducible benchmark for freeway traffic reconstruction, with demonstrated use cases across multiple independent freeway sites.
\item This paper is fully computational reproducible. Data and code are publicly available at \url{https://github.com/trafficwaves/ASMx}.
\end{enumerate}

This article, together with the accompanying code repository, can also serve as a tutorial to help colleagues and students understand both computational methods and freeway systems, providing a ground for the development of improved future methods and bringing these approaches closer to real-world large-scale implementation.
The reminder of the paper is organized as follows. Section~\ref{sec:method} details the ASM method and its convolutional implementation. Section~\ref{sec:results} presents the calibration results, illustrates how the parameters evolve over iteration. Section~\ref{sec:discussions} discusses the convexity and reproducibility of the calibration problem, demonstrates the use cases of the calibrated model across multiple locations, and future extensions for ASM. Finally, Section~\ref{sec:conclusion} concludes the paper and discusses the lessons learned from this implementation for the reproducible research in transportation research.
\section{Methods}
\label{sec:method}
\subsection{Adaptive Smoothing Method}
The original Adaptive Smoothing Method (ASM) \cite{treiber2003adaptive,treiber2011reconstructing} is formulated in a continuous space-time framework. Let ${\vec{\Theta}} = \{ v_{\text{cong}}, v_{\text{free}}, \delta, \tau, V_{\text{thr}}, \Delta V \}$ denotes the set of parameters for the ASM model, where $v_{\text{cong}}$ is the wave speed in congested traffic, $v_{\text{free}}$ is the wave speed in free-flowing traffic, $\delta$ defines the spatial smoothing width, $\tau$ defines the temporal smoothing width, $V_{\text{thr}}$ specifies the threshold velocity marking the transition from congested to free traffic, and $\Delta V$ characterizes the width of this transition. For practical applications, it can be written in a matrix format \cite{chen2024forecasting}, treating it as a matrix convolutional operation \cite{schreiter2010two}. 

Let $\mathcal{\tilde{X}} \in \mathbb{R}^{K \times N}$ with columns $\vec{\tilde x}_1, \dots , \vec{\tilde x}_K \in \mathbb{R}^{N}$ represent the traffic speed measurements collected from $N$ road sensors with $K$ time steps. Let $\mathcal{X} \in \mathbb{R}^{S \times T}$ represent the high-resolution target data to be reconstructed, where $S \geq N$ (spatially higher resolution) and $T \geq K$ (temporally higher resolution). $\Omega$ denotes the index set of observed entries in $\mathcal{X}$. 

To connect the sparse observed data $\mathcal{\tilde{X}}$ to the higher-resolution data $\mathcal{X}$, define the observed index set $\Omega$ explicitly in terms of sensor and time-step indices. Assume that the observed spatial locations correspond to sensor indices $\{s_1, s_2, \dots, s_N\}\subseteq\{1,\dots,S\}$ and the observed time points correspond to indices $\{t_1, t_2, \dots, t_K\}\subseteq\{1,\dots,T\}$. The observed index set is defined explicitly as:

\begin{equation}
    \Omega = \{(s_n, t_k)\ |\ n = 1,\dots,N;\ k = 1,\dots,K\}.
\end{equation}

We define the projection $P_{\Omega}$ of the sparse measurements $\mathcal{\tilde{X}}$ onto the high-resolution matrix $\mathcal{X}$ as follows:

\begin{equation}
P_{\Omega}(\mathcal{X})_{ij}
=
\begin{cases}
\mathcal{X}_{ij}, & (i,j)\in\Omega,\\
0, & \text{otherwise},
\end{cases}  
\end{equation}

The relationship between the observed data matrix $\mathcal{\tilde{X}}$ and the projected high-resolution matrix $P_{\Omega}(\mathcal{X})$ is explicitly given by:

\begin{equation}
\mathcal{X}_{s_n t_k} = \mathcal{\tilde{X}}_{kn}, \quad\text{for all } k = 1,\dots,K,\; n = 1,\dots,N.
\end{equation}

Similarly, a mask matrix $\mathcal{M} \in \mathbb{R}^{S \times T}$ can be defined as

\begin{equation}
\mathcal{M}_{ij}
=
\begin{cases}
1, & (i,j)\in\Omega,\\
0, & \text{otherwise}.
\end{cases}  
\end{equation}

In the ASM, the sparse spatio-temporal data matrix $P_\Omega(\mathcal{X})$  is processed using convolution with two kernels: $\mathcal{K}_c \in \mathbb{R}^{(2S+1) \times (2T+1)}$, tuned for congested wave propagation, and $\mathcal{K}_f \in \mathbb{R}^{(2S+1) \times (2T+1)}$, tuned for free-flow waves. An element at grid position $(p,q)$ effectively corresponds to the spatial offset $x_p = p \times \Delta S$ and the time offset $t_q = q \times \Delta T$, where $p \in \{-S, \dots, 0, \dots, S\}$ and $q \in \{-T, \dots, 0, \dots, T\}$. The fixed anisotropic kernels are based on empirical traffic flow characteristics observed in prior studies. The physical meaning of the anisotropic kernels is that traffic disturbances (or signals \cite{coifman2023lwr,coifman2024microscopic}) propagate differently in the upstream and downstream directions. These empirical patterns have been validated in multiple studies \cite{treiber2013traffic,coifman2023lwr,ji2024scalable}.

\begin{align}
\mathcal{K}_c(p, q; \vec{\Theta}) &= \exp\left(-\frac{\left| t_q - \frac{x_p}{c_{\text{cong}}} \right|}{\tau} - \frac{\left| x_p \right|}{\delta} \right), 
\label{eq:kernel1}
\\
\mathcal{K}_f(p, q; \vec{\Theta}) &= \exp\left(-\frac{\left| t_q - \frac{x_p}{c_{\text{free}}} \right|}{\tau} - \frac{\left| x_p \right|}{\delta} \right).
\label{eq:kernel2}
\end{align}

Note that the kernels in \eqref{eq:kernel1} and \eqref{eq:kernel2} are assumed to be stationary, meaning its shape and parameters do not change over time or space. We then form the two raw reconstructions by convolution:

\begin{equation}
\mathcal{Z}_c \;=\; \frac{\mathcal{K}_c(\vec{\Theta})\;*\;P_\Omega(\mathcal{X})}{\mathcal{K}_c(\vec{\Theta})\;*\;\mathcal M}, 
\qquad
\mathcal{Z}_f \;=\; \frac{\mathcal{K}_f(\vec{\Theta})\;*\;P_\Omega(\mathcal{X})}{\mathcal{K}_f(\vec{\Theta})\;*\;\mathcal M}, 
\end{equation}
where the operator $*$ denotes the convolution operation. With the two estimates computed, a weight matrix $\mathcal{W}$, as a nonlinear adaptive filter, is introduced to enable the adaptive selection between them, and is defined as follows:

\begin{equation}
\mathcal{W}(\mathcal{Z}_c,\mathcal{Z}_f;\vec{\Theta}) \;=\; \frac12\Bigl[\,1+\tanh\!\bigl(\,\frac{(V_{\rm thr}-\min(\mathcal{Z}_c,\mathcal{Z}_f))}{{\Delta V}}\bigr)\Bigr],
\label{eq:nonconvex}
\end{equation}
and the reconstructed speed matrix $\mathcal{X}$ is computed by combining the congested and free-flow estimates with the adaptive blending weights:

\begin{equation}
\mathcal{X} \;=\; \mathcal{W}\,\mathcal{Z}_c \;+\;(1-\mathcal{W})\,\mathcal{Z}_f.
\end{equation}
The nonlinear adaptive speed filter is designed to smoothly transition between free-flow and congested traffic regimes with an s-shaped nonlinear function. A detailed justification of this design can be found in \appendixname~\ref{appendix:filter}.

\subsection{Calibration}
Mobility observation instruments \cite{seo2020evaluation, gloudemans202324, schicktanz2025dlr}, such as the I-24 MOTION testbed \cite{gloudemans202324}, enable the collection of ground truth mean speed data $\mathcal{\hat X}$ for calibrating the parameter set $\vec{\Theta}$ in ASM. The calibration objective is to minimize the loss function $\mathcal{L}$, with the optimization problem formulated as
\begin{equation}
\vec{\Theta}^* =  \mathop{\arg\!\min}_{\vec{\Theta}} \mathcal{L}(\mathcal{X}(\mathcal{\tilde{X}};\vec{\Theta}), \mathcal{\hat X}),
\label{eq:calibration}
\end{equation}
where $\vec{\Theta}$ denotes the parameters being optimized and $\mathcal{\tilde{X}}$ denotes the sparse sensor observation.
The loss function $\mathcal{L}$ in~\eqref{eq:calibration} is chosen as the weighted root-mean-square deviation (WRMSE) between the reconstructed speed matrix $\mathcal{X}(\mathcal{\tilde{X}};\vec{\Theta})$ and the ground truth $\mathcal{\hat X}$, given by

\begin{equation}
\mathcal{L}(\mathcal{X}(\mathcal{\tilde{X}};\vec{\Theta}), \mathcal{\hat X}) = \sqrt{\frac{1}{|\Omega|}\sum_{(i,j)\in\Omega} w_{ij} \left(\mathcal{X}_{ij}(\mathcal{\tilde{X}};\vec{\Theta}) - \mathcal{\hat X}_{ij}\right)^2},
\label{eq:loss}
\end{equation}
where $w_{ij}$ is the weight for each entry in the loss function, which can be set to $w$ for $\mathcal{\hat X}_{ij} \leq v_c$, otherwise set to $1$ for $\mathcal{\hat X}_{ij} > v_c$, where $v_c$ is a threshold speed. The choice of this weighted loss function is to better handle the low-speed data, which is the traffic waves in the freeway operation context \cite{ji2025stop}. In our experiments, we set $w = 10$ and $v_c = 24.14$ km/h. The learning process stops after a fixed number of epochs (in this study, N = 1000). We track the best validation WRMSE for the calibrated parameters during training.

\subsection{Implementation}
Based on the matrix formulation of ASM, we implement the method using PyTorch \cite{paszke2019pytorch}. Following the efficient implementation via Fast Fourier Transform (FFT) proposed by \cite{schreiter2010two}, we replace the standard convolution operation ($*$) in PyTorch with an FFT-based approach. A comparison of the computational complexities is presented in Table~\ref{tab:complexity}. The complexity has dropped from quadratic to quasi-linear. 

\begin{table}[htbp]
\footnotesize
\centering
\caption{A comparison of the computational complexity between the FFT and the standard convolution operation in PyTorch, assuming the matrix has dimensions $\mathbb{R}^{S \times T}$.}
\begin{tabular}{lcc}
\toprule
Operation & PyTorch call & Complexity \\
\midrule
Hadamard products      & \texttt{*}      & \(\mathcal O(S^2T^2)\)       \\
Real FFT (inputs \& kernels) & \texttt{rfftn}  & \(\mathcal O(ST\log(ST))\)\\
Inverse real FFT       & \texttt{irfftn} & \(\mathcal O(ST\log(ST))\)\\
\bottomrule
\end{tabular}
\label{tab:complexity}
\end{table}

All operations, including FFT, slicing, division, and the $\tanh$ activation, are differentiable and implemented in PyTorch, making the parameters trainable and allowing integration with other deep neural networks built in PyTorch.
The model parameters are optimized using the \texttt{Adam} optimizer \cite{kingma2014adam} to solve the calibration problem in~\eqref{eq:calibration}, with a learning rate set to 0.01. At each iteration, parameter values are rounded to two decimal places for ease of application. Additionally, the free-flow speed parameter $v_{\text{free}}$ is constrained to not exceed 96.56 km/h (60 mph) using a clamping operation in PyTorch.

\subsection{Computational reproducibility report}

All experiments conducted in this article are performed on a machine with two NVIDIA RTX A6000 GPUs, AMD Ryzen Threadripper 3960X 24-Core Processor and with 256 GB of RAM. 

The GPU time and CPU time reported in this article are operated in this hardware environment. Hardware-independent metrics, including FLOPs (Floating Point Operations), MACs (Multiply-Accumulate operations), throughput, and memory footprint are also reported providing a more robust and hardware-independent evaluation of the implementation's performance. FLOPs and MACs are estimated by Python package \texttt{thop}~\cite{thop_pypi}.

In our experiments, we enforce fixed seeds (seed=42) for \texttt{numpy}, \texttt{torch}, and \texttt{CUDA} operations at the start of execution. This practice ensures that the results are reproducible across different runs and environments.

\section{Results}
\label{sec:results}
\subsection{Data preprocessing}
I-24 MOTION \cite{gloudemans202324} is a traffic instrument located along 6.76 km of interstate roadway near Nashville, Tennessee.
I-24 SMART Corridor is a 27.36 km corridor between Nashville and Murfreesboro. The I-24 Radar Detector System (RDS) is a subsystem equipped with recently calibrated Wavetronix HD sensors \cite{wavetronix_smartsensor_hd}, that provides detailed 30-second aggregated speed, occupancy and volume. The two systems have spatio-temporal overlap, where I-24 MOTION generates the ground truth $\mathcal{\hat{X}}$ and I-24 RDS generates the sparse sensor observation $\mathcal{\tilde{X}}$, making the calibration problem in~\eqref{eq:calibration} possible. I-24 MOTION data is processed to a spatio-temporal resolution of 4 seconds and 32 meters (approximately 0.02 miles) \cite{ji2024virtual}. I-24 RDS data is sampled at every 30 seconds, with sensors spaced at an average interval of 483 meters (approximately 0.3 miles). 
\begin{figure}[htbp]
    \centering
    \begin{minipage}{0.8\linewidth}
        \centering
        \includegraphics[width=\linewidth]{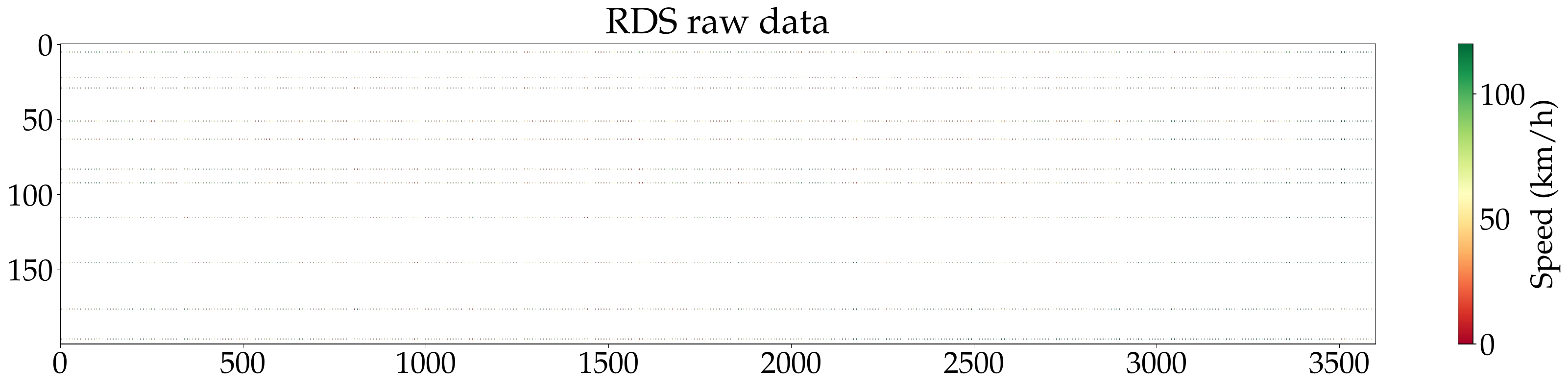}
        \subcaption{Preprocessed sparse speed matrix}
    \end{minipage}
    
    \begin{minipage}{0.8\linewidth}
        \centering
        \includegraphics[width=\linewidth]{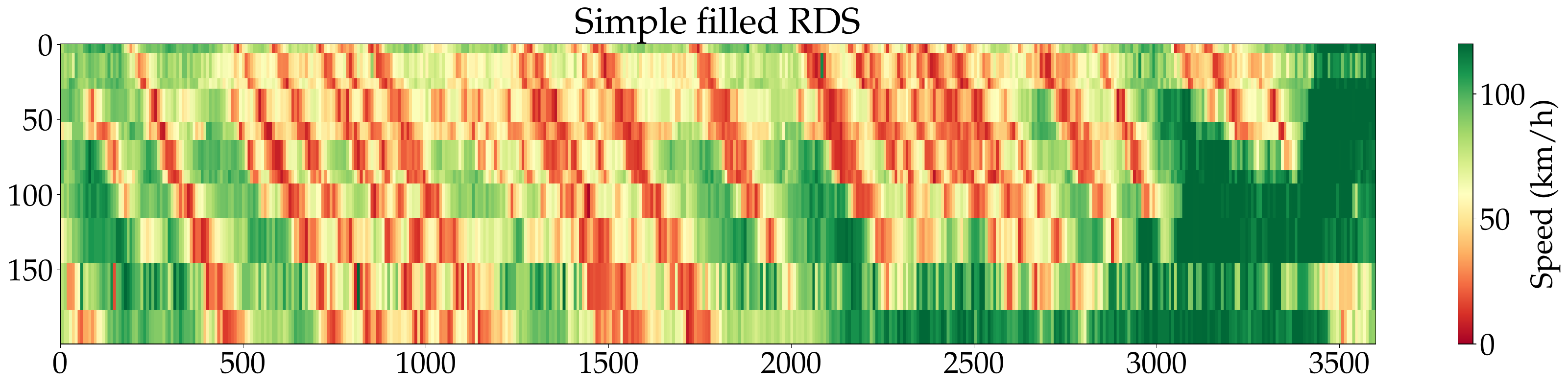}
        \subcaption{Simple filled speed matrix}
        \label{fig:ffill}
    \end{minipage}

    \begin{minipage}{0.8\linewidth}
        \centering
        \includegraphics[width=\linewidth]{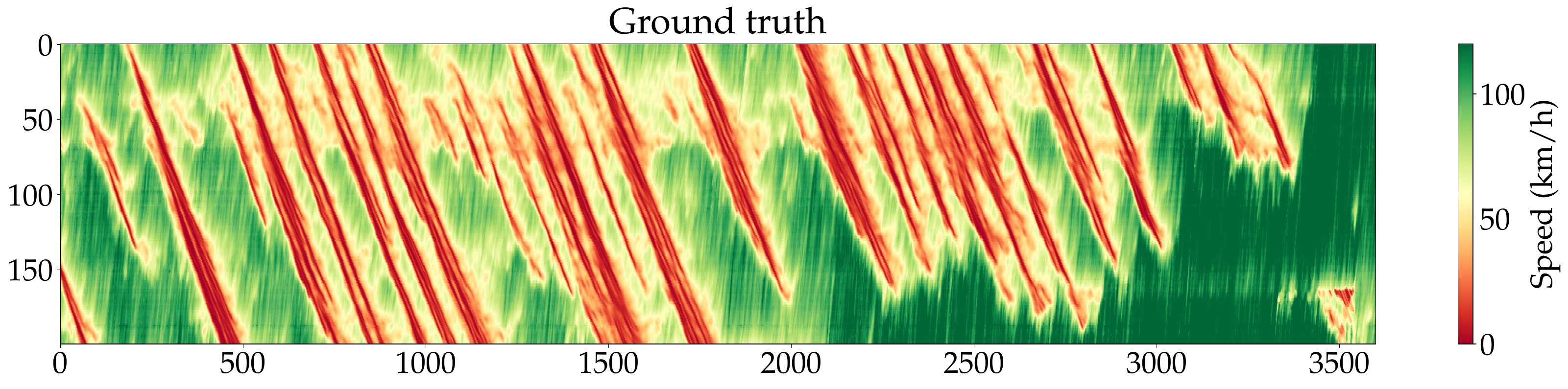}
        \subcaption{Ground truth speed matrix}
    \end{minipage}
        \caption{Comparison between the preprocessed sparse speed matrix and the ground-truth speed matrix for Interstate 24, spanning Mile Marker 58.7 to Mile Marker 62.7. The x-axis represents the temporal index, while the y-axis corresponds to the spatial index; each matrix entry is color-coded according to speed. Both matrices are discretized at 4-second temporal intervals and 32-meter spatial intervals. Figure~\ref{fig:ffill} illustrates the  simple filled speed matrix, in which missing values are imputed using the most recently observed measurement in time and space, consistent with standard practices in traffic management center operations.}
    \label{fig:comparison}
\end{figure}
I-24 MOTION speed data is processed following Edie's definition \cite{edie1963discussion} at a spatial resolution of 32 meters and temporal resolution of 4 seconds. To accurately compute Edie's mean speed, we employ the parallelogram method that incorporates traffic wave speed, consistent with prior practices in \cite{knoop2012quantifying,tsanakas2022generating,ji2024virtual}.
The RDS data is then aligned to this resolution by matching to the nearest values. In this article, we used data collected on July 9th, 2024, from 6:00 AM to 10:00 AM on westbound Interstate 24 to calibrate the ASM. A visualization of the raw RDS data and the MOTION ground truth is shown in Figure~\ref{fig:comparison}.
\subsection{Calibration results}
The initial guess we set for the calibration is based on \cite{treiber2011reconstructing}. The calibration process on lane 1 data using \texttt{Adam} is performed with 1000 epochs, tracking the parameter evolution in Figure~\ref{fig:params_iteration} and the loss function WRMSE in Figure~\ref{fig:rmse_iteration}. All the logs from the experiments, including the calibration time, the parameter iterations, are recorded in the open code repository under the directory \texttt{logs/calibration}. Results presented in this article are from the experiment run ID \texttt{20250607\_221107}.

\begin{figure}[htbp]
    \centering
    \includegraphics[width=0.95\linewidth]{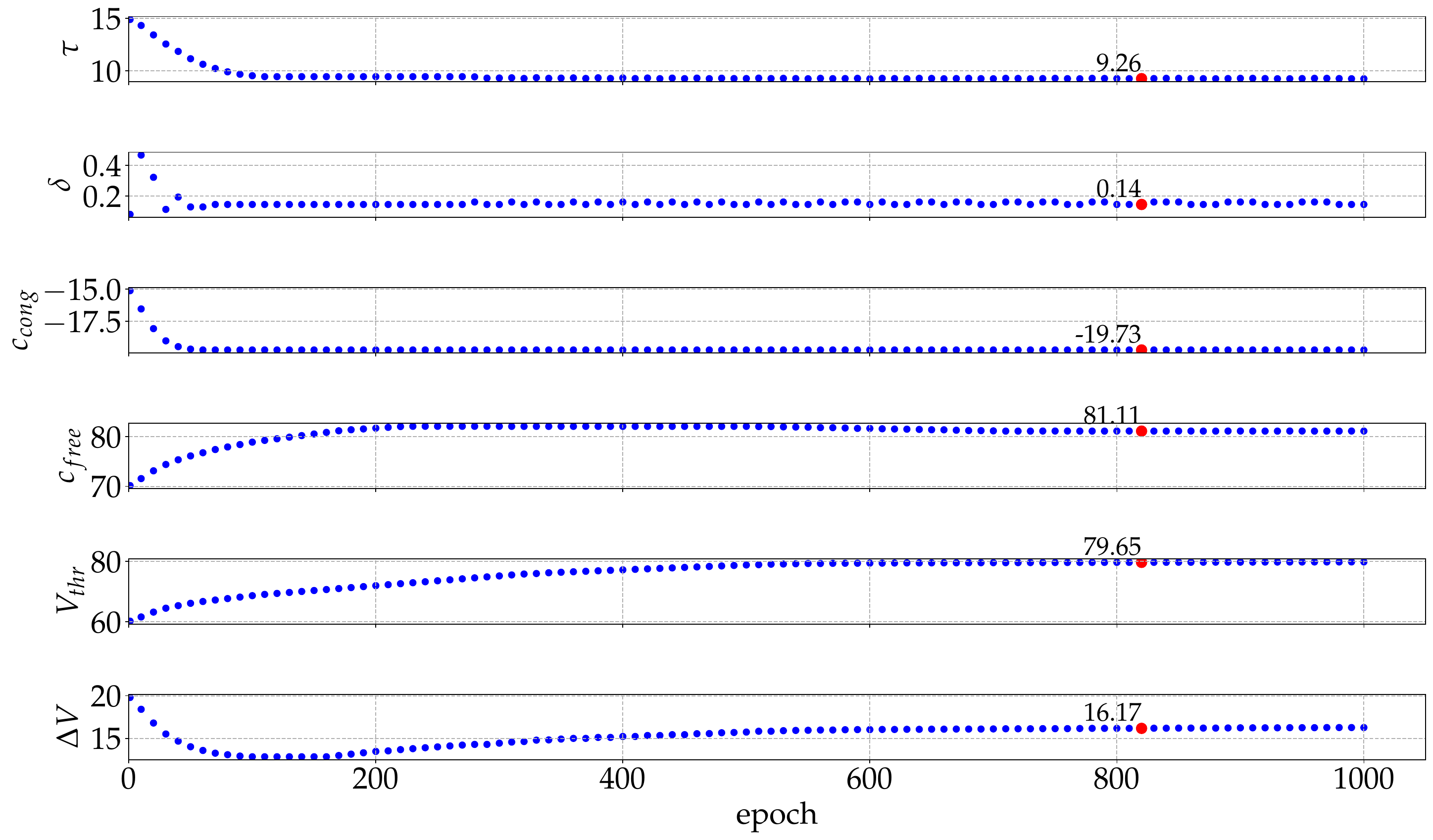}
    \caption{Parameter evolution over epochs for the lane 1 calibration task ($\tau$, $\delta$, $c_\text{cong}$, $c_\text{free}$, $V_\text{thr}$, and $\Delta V$ from top to bottom). The starting point is the initial guess suggested by \cite{treiber2011reconstructing}, with the red dot indicating the best parameter values achieved during the calibration process.}
    \label{fig:params_iteration}
\end{figure}

As shown in Figure~\ref{fig:params_iteration}, the parameters converge to a stable value after approximately 600 epochs. The final calibrated parameters for each lane are summarized in Table~\ref{tab:params}. In this article, lane 1 refer to the leftmost lane, lane 2 refers to the second leftmost lane, and so on. 
The time taken for the calibration process is approximately 50 seconds per lane with GPU, with the total time for all four lanes being around 4 minutes. As shown in Figure~\ref{fig:rmse_iteration}, the loss function converges fast to a stable value after approximately 50 epochs.

\begin{figure}[htbp]
    \centering
    \includegraphics[width=0.95\linewidth]{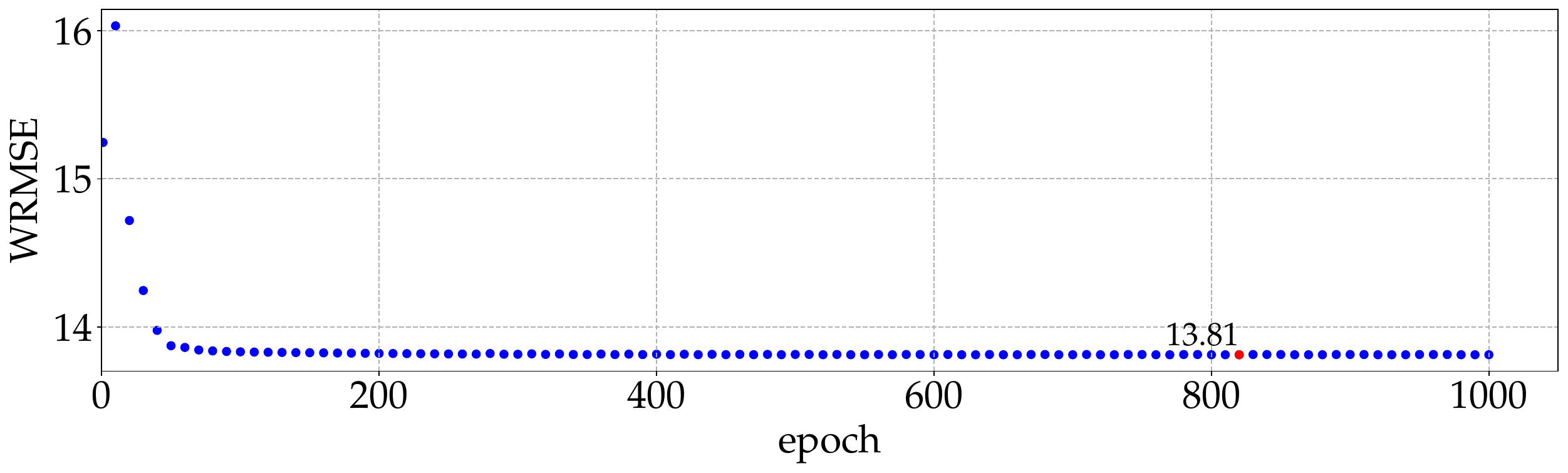}
    \caption{The loss function over the calibration epochs for the lane 1 calibration task, with the red dot indicating the best result achieved during the process.}
    \label{fig:rmse_iteration}
\end{figure}

\begin{table}[htbp]
\centering
\footnotesize
\caption{Calibrated parameters for each lane. The initial guess is based on the parameters suggested by \cite{treiber2011reconstructing}. The time taken for each calibration is reported in seconds.}
\label{tab:params}
\begin{tabular}{ccccccc}
\hline
 Parameter & Unit& Initial Guess& Lane 1 & Lane 2 & Lane 3 & Lane 4 \\ \hline
$\tau$ & seconds& 15.00 & 9.27 & 7.86 & 8.25 & 11.56 \\
$\delta$ & km& 0.24  & 0.14 & 0.14 & 0.14 & 0.16 \\
$c_\text{cong}$ & km/h&-15.00 & -19.73 & -20.20 & -21.02 & -20.57 \\
$c_\text{free}$ & km/h&70.00      & 81.11 & 80.13 & 81.43 & 96.56 \\
$V_{\text{thr}}$ & km/h &60.00  & 79.78 & 67.59 & 65.92 & 59.42 \\
$\Delta V$    & km/h  &20.00 & 16.27 & 12.39 & 13.18 & 14.20  \\ \hline
Time (GPU)   & seconds & - & 49.07  &48.07   & 48.01 &  48.16 \\ \hline
\end{tabular}
\end{table}

As shown in Table~\ref{tab:params}, the calibrated values of $c_\text{cong}$ range from 19 km/h to 21 km/h across different lanes, which is consistent with values reported in previous studies~\cite{ji2024scalable}. The initial guesses for $\tau$ and $\delta$ were set to half the temporal sampling interval and the average spatial interval of the sensors, respectively (i.e., 15 seconds and 0.24 km), as recommended by~\cite{treiber2011reconstructing}. The calibrated values for $\tau$ and $\delta$ are generally lower than these initial guesses, suggesting that optimal values are typically about one third of the corresponding intervals. The difference in the calibrated values of $V_{\text{thr}}$ and $\Delta V$ across lanes indicates that the traffic conditions vary significantly, with lane 1 (the high occupancy vehicle lane) having a higher threshold speed and a wider transition width compared to lane 4 (the rightmost lane). Figure~\ref{fig:asm-comparison} and Figure~\ref{fig:lane4} show the comparison of the reconstructed speed matrix using calibrated ASM parameters and the ground truth speed matrix from I-24 MOTION for lane 1 and lane 4, respectively.

\begin{figure}[htbp]
    \centering
    \begin{minipage}{0.80\linewidth}
        \centering
        \includegraphics[width=\linewidth]{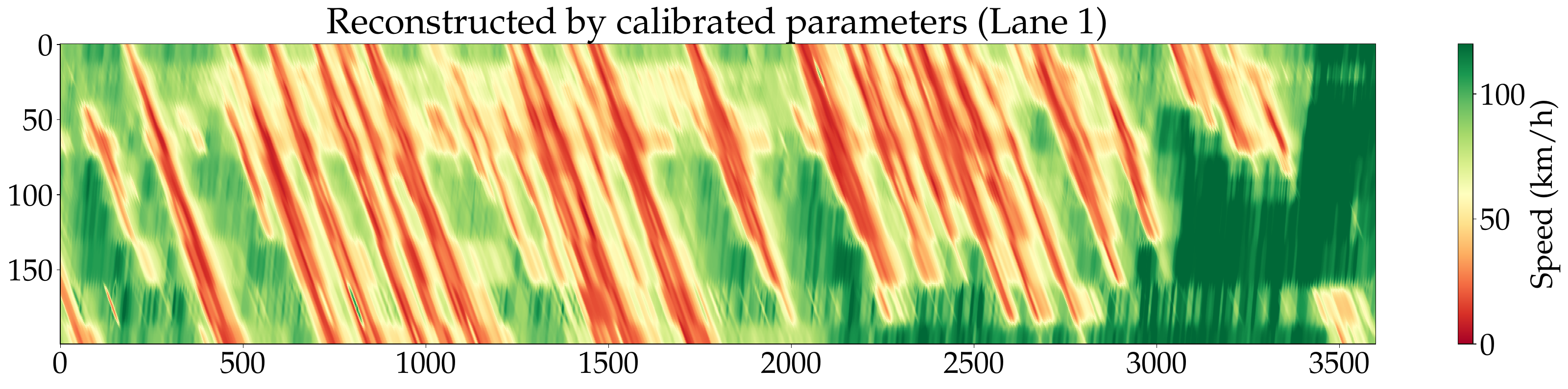}
        \subcaption{Reconstructed speed matrix with calibrated parameters on lane 1}
    \end{minipage}

    \begin{minipage}{0.80\linewidth}
        \centering
        \includegraphics[width=\linewidth]{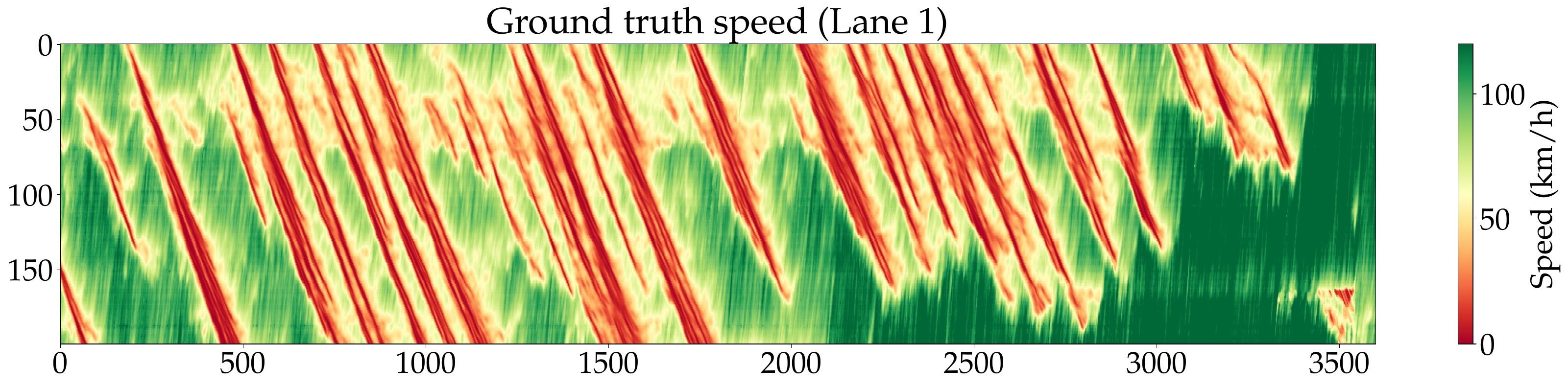}
        \subcaption{Ground truth speed matrix on lane 1}
    \end{minipage}
    \caption{Comparison of the reconstructed speed matrix using calibrated ASM parameters and the ground truth speed matrix from I-24 MOTION. The data shown in this figure is from lane 1 (lane 1 refers to the left most lane), July 9th, 2024, from 6:00 AM to 10:00 AM. Lane 1 is also the high occupancy vehicle (HOV) lane, which is typically used by sedans.}
    \label{fig:asm-comparison}
\end{figure}

The results indicate that lane 1 exhibits greater speed variability and more pronounced traveling wave patterns under free-flow conditions compared to lane 4. In contrast, lane 4 shows less distinct spatio-temporal structure in the free-flow regime, with maximum speeds seldom reaching the posted speed limit of 113 km/h. This likely explains why the calibrated value of $c_\text{free}$ for lane 4 reaches the imposed upper bound of 96.56 km/h, as the spatio-temporal correlation at higher speeds is weaker.

\begin{figure}[htbp]
    \centering
    \begin{minipage}{0.80\linewidth}
        \centering
        \includegraphics[width=\linewidth]{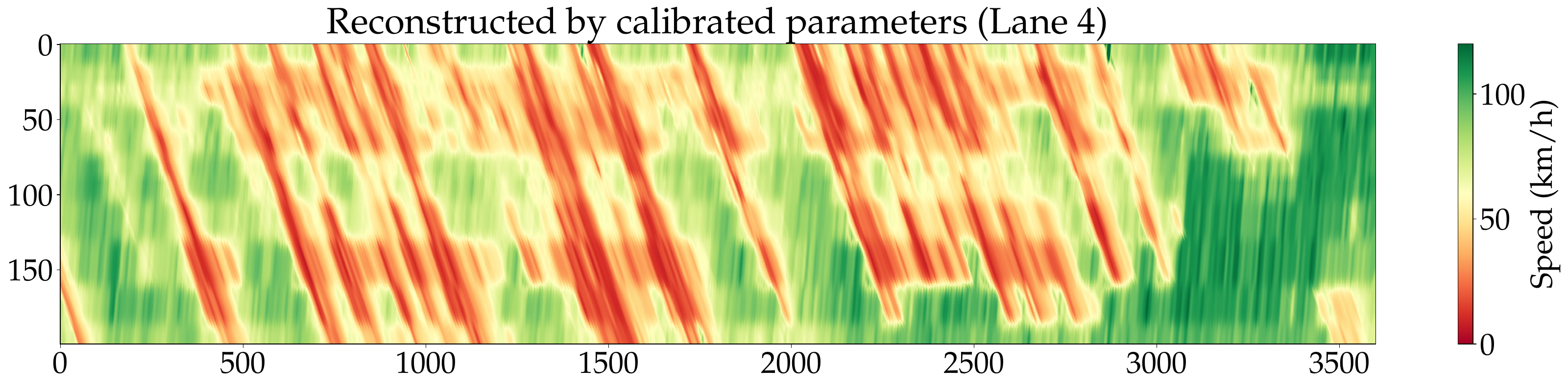}
        \subcaption{Reconstructed speed matrix with calibrated parameters on lane 4}
    \end{minipage}
    \\
    \begin{minipage}{0.80\linewidth}
        \centering
        \includegraphics[width=\linewidth]{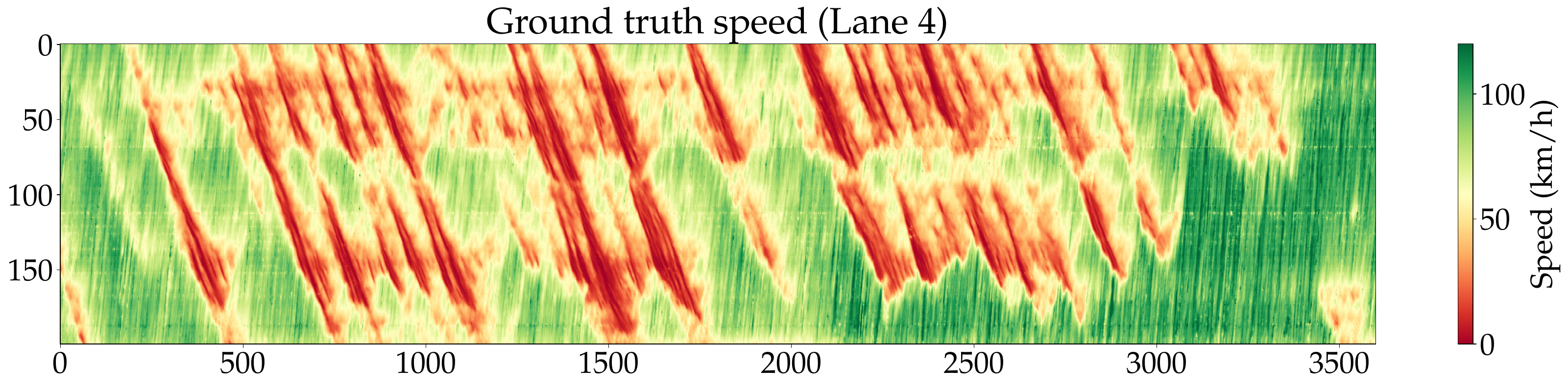}
        \subcaption{Ground truth speed matrix on lane 4}
    \end{minipage}
    \\
    \caption{Comparison of the reconstructed speed matrix using calibrated ASM parameters and the ground truth speed matrix from I-24 MOTION. The data shown in this figure is from lane 4 (lane 4 refers to the right most lane), July 9th, 2024, from 6:00 AM to 10:00 AM. Lane 4 is primarily used by a mix of trucks and sedans and is adjacent to on-ramps and off-ramps.}
    \label{fig:lane4}
\end{figure}
\subsection{Evaluation}
Figure~\ref{fig:rmse_iteration} demonstrates that the loss function converges rapidly to a stable value, indicating the attainable accuracy of ASM given the current input data. This suggests that further improvements in reconstruction quality would require either higher-quality sensor data or methodological advances beyond ASM. To provide a comprehensive evaluation and establish benchmark criteria beyond the commonly used Root Mean Square Error (RMSE) for future methods, we further evaluate the reconstruction error in terms of speed distribution, spatio-temporal distribution, spatial distribution, and the intersection-over-union (IoU) of traffic waves.

\subsubsection{Speed distribution}
Figure~\ref{fig:speed_histogram} compares the speed distributions for the ground truth, the original ASM parameters, and the calibrated ASM parameters. The calibrated ASM parameters show better alignment with the low-speed region, but a distribution mismatch remains, as indicated by the directional shift in the figure. To quantify this, we compute the Wasserstein distance \cite{villani2008optimal} between the ground truth and both ASM reconstructions using \texttt{scipy.stats.wasserstein\_distance}. The Wasserstein distance for the original ASM is 5.09 km/h, while for the calibrated ASM it is 3.46 km/h, representing a 31.96\% improvement. For reference, the RMSE improves from 11.70 km/h (original ASM) to 11.41 km/h (calibrated ASM), a 2.48\% reduction. The Wasserstein distance is a recommended metric for the evaluation of traffic speed reconstruction tasks, as it better captures the distributional differences compared to RMSE.

\begin{figure}[htbp]
    \centering
    \includegraphics[width=0.95\linewidth]{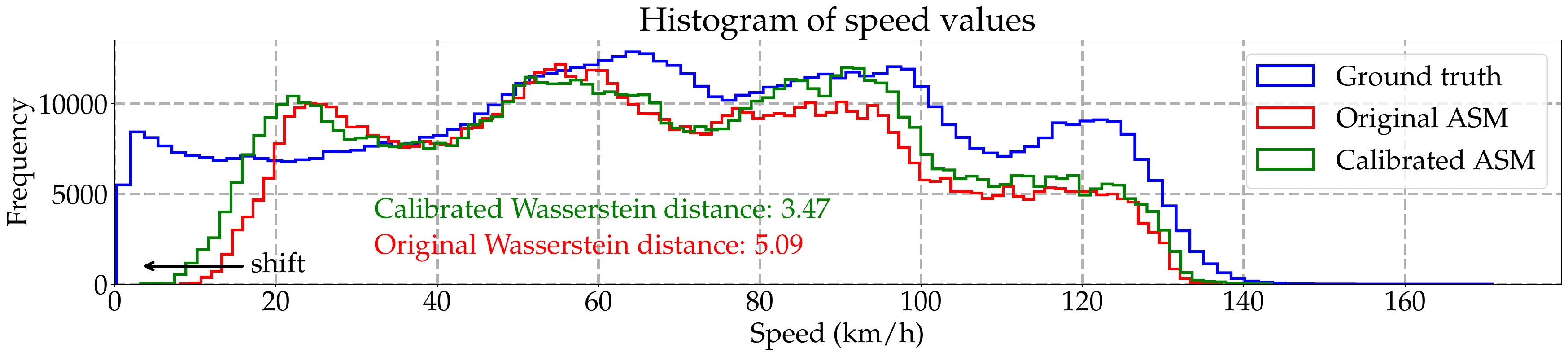}
    \caption{Histogram of the speed distribution for the ground truth, reconstruction from the original ASM parameters, and the calibrated ASM parameters. The data shown in this figure is from lane 1 (lane 1 refers to the left most lane), July 9th, 2024, from 6:00 AM to 10:00 AM.}
    \label{fig:speed_histogram}
\end{figure}

\subsubsection{Speed error spatio-temporal distribution}
To investigate the reconstruction error distribution, Figure~\ref{fig:error} shows the error distribution on the space-time diagram for each point in space and time. 
\begin{figure}[htbp]
    \centering
    \includegraphics[width=0.95\linewidth]{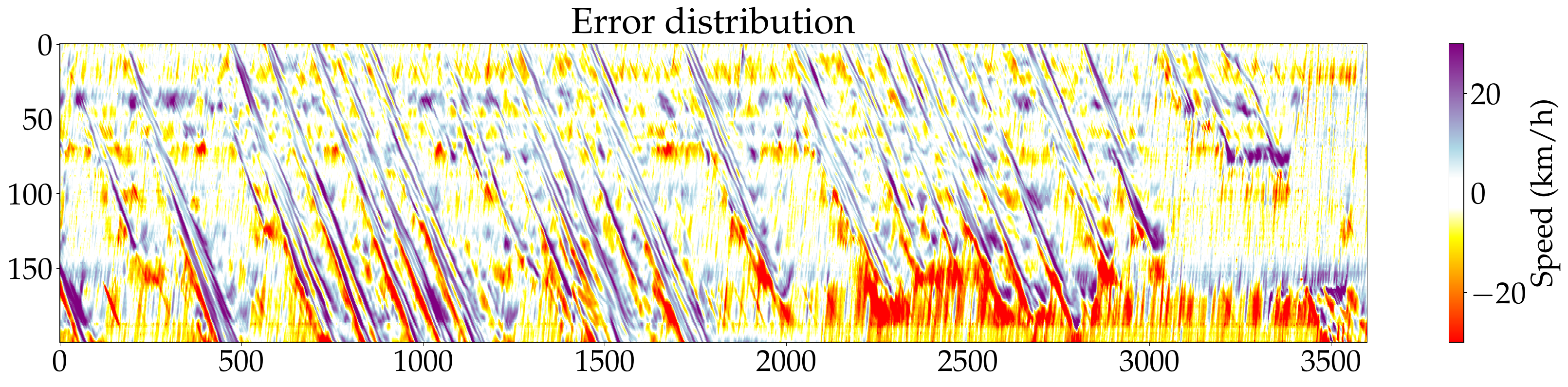}
    \caption{The error distribution of the ASM-calibrated parameters in the speed field, relative to the ground truth, where ``+" indicates the actual speed is higher and ``–" indicates the actual speed is lower compared to the ground truth.}
    \label{fig:error}
\end{figure}

As illustrated in Figure~\ref{fig:error}, the reconstruction error under free-flow conditions is generally small, with most significant errors concentrated within regions exhibiting traffic waves. These errors can be attributed to two primary sources: (\romannumeral1) aggregation bias inherent to the sensor measurements \cite{treiber2013traffic}, which arises from discrepancies between aggregated and actual speeds (as shown in \figurename~\ref{fig:residuals} top left area); and (\romannumeral2) limitations of the ASM model itself, as it is a kernel-based smoothing approach that does not incorporate additional traffic variables such as density. Consequently, ASM tends to overestimate wave propagation ((as shown in \figurename~\ref{fig:residuals} bottom middle area)), as evidenced by the predominance of underestimated speeds (i.e., the red regions) in areas where traffic waves dissipate.

\begin{figure}[htbp]
    \centering
    \includegraphics[width=0.9\linewidth]{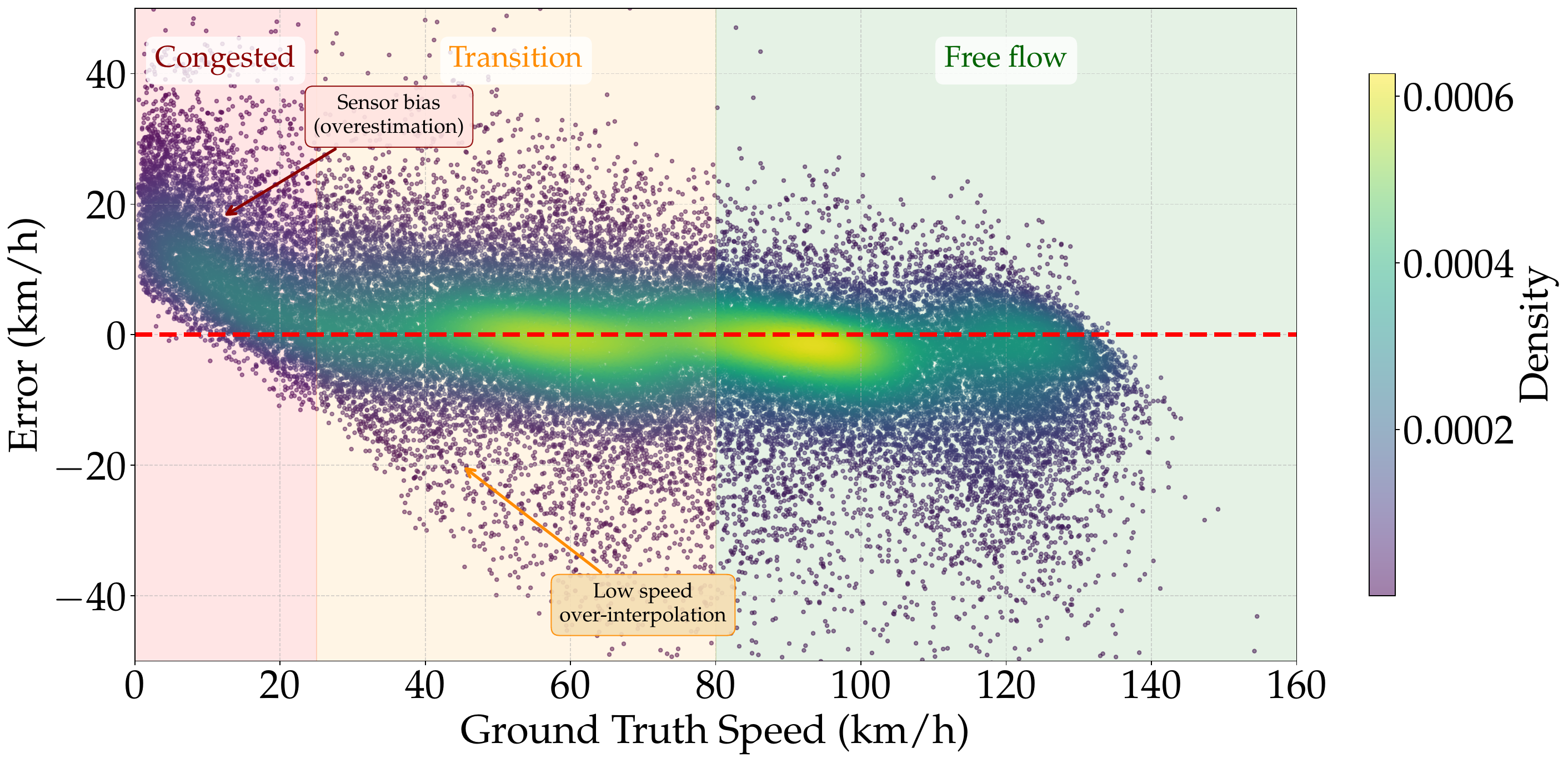}
    \caption{Residual analysis between calibrated ASM and ground truth. Scatter-density plot of speed error versus ground-truth speed, with color indicating point density. Three traffic regimes are highlighted: congested (low speeds), transition, and free flow (high speeds). Data from lane 1 on July 9, 2024.}
    \label{fig:residuals}
\end{figure}

\begin{table}[ht]
\centering
\footnotesize
\caption{Intersection-over-Union (IoU) of traffic waves at different speed thresholds $v_c$ (km/h). ``Only in Reconstruction (OiREC)'' and ``Only in Ground Truth (OiGT)'' denote the proportion of wave regions identified exclusively by the ASM reconstruction and the ground truth, respectively.}
\label{tab:iou}
\begin{tabular}{ccccccc}
\hline
Threshold & IoU (O) & IoU (C) & OiREC (O) & OiREC (C) & OiGT (O) & OiGT (C) \\
\midrule
8 & 0.0000 & \underline{0.0071} & 0.0000 & \underline{0.0007} & 1.0000 & \underline{0.9922}\\
16& 0.0850 & \underline{0.2011} & 0.0096 & \underline{0.0442} & 0.9055 & \underline{0.7547} \\
24& 0.4167 & \underline{0.5279} & 0.0638 & \underline{0.1043} & 0.5195 & \underline{0.3678} \\
32& 0.6561 & \underline{0.6869} & 0.0857 & \underline{0.1100} & 0.2583 & \underline{0.2031} \\
40& 0.7541 & \underline{0.7649} & 0.0869 & \underline{0.0997} & 0.1589 & \underline{0.1354} \\
48& 0.7859 & \underline{0.7877} & 0.1012 & \underline{0.1042} & 0.1130 & \underline{0.1081} \\
\hline
\end{tabular}
\end{table}

To further evaluate reconstruction quality, we evaluate the intersection-over-union (IoU) of traffic waves, defined as the ratio of the intersection area to the union area between regions below a critical speed threshold $v_c$ in both the reconstruction and ground truth. The columns ``Only in Reconstruction (OiREC)'' and ``Only in Ground Truth (OiGT)'' indicate the proportions of wave regions detected exclusively by the ASM reconstruction (false positives) and exclusively by the ground truth (false negatives), respectively. Table~\ref{tab:iou} reports IoU values for several speed thresholds $v_c$. Higher IoU and lower OiREC/OiGT values indicate better agreement between the ASM reconstruction and ground truth. For example, at $v_c = 24$ km/h, the IoU improves from 0.4167 (original ASM) to 0.5279 (calibrated ASM), a 26.6\% increase. However, the OiREC also increases, indicating that the calibrated ASM introduces more false positives, corresponding to the red regions described in Figure~\ref{fig:error}. These can be used as metrics to quantify the hierarchical speed reconstruction error, especially for the low-speed traffic waves, which are critical for freeway operation and control.

\subsubsection{Spatial distribution of reconstruction error}
\begin{figure}[htbp]
    \centering
    \includegraphics[width=0.95\linewidth]{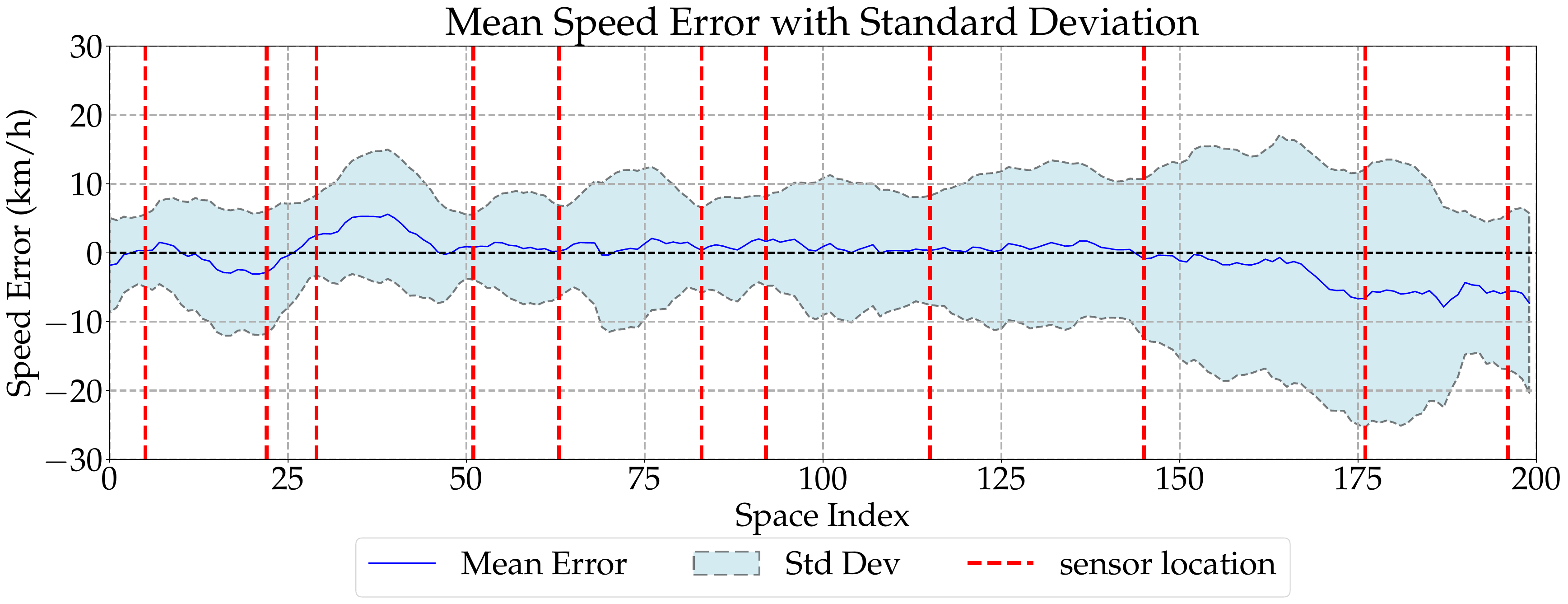}
    \caption{Mean and standard deviation of the reconstruction error by spatial location. The x-axis shows the spatial index, while the y-axis indicates the mean error at each location, with shaded regions representing one standard deviation. Red dashed lines indicate the locations of the sensors. Data shown is for lane 1 (leftmost lane), July 9th, 2024, from 6:00 AM to 10:00 AM.}
    \label{fig:mean_error_location}
\end{figure}

We further analyze the spatial distribution of reconstruction error in Figure~\ref{fig:mean_error_location}. For each spatial location, we compute the mean error and its standard deviation, with the shaded region representing one standard deviation. Sensor locations are indicated by red dashed lines. The results show that both the magnitude and variability of reconstruction error are generally lower near sensor locations, while larger errors and greater variability occur between sensors. Notably, the sensor at index 176 exhibits a higher mean error, which may be due to sensor-specific biases such as sensor configuration issues, spatial and temporal misalignment. This evaluation provides insight into sensor data quality, helps identify potential sensor issues, and can inform improvements to the sensor calibration process.

\subsection{Cross validation}
Parameters calibrated using data from July 9, 2024, were applied to validate the model against four additional days (July 8, 10, 11, and 12). The performance is evaluated using two metrics: Root Mean Square Error (RMSE) and Wasserstein distance, both measured in km/h. The RMSE and Wasserstein distance results are shown in \figurename~\ref{fig:cross-validation-compared}. As shown in\figurename~\ref{fig:cross-validation-compared}, the model maintains consistent performance across different days and lanes, with the RMSE generally ranging between 8.23 and 12.53 km/h and the Wasserstein distance between 1.28 and 4.20 km/h. This indicates that the calibrated parameters can be effectively applied to reconstruct traffic speeds on different days without the need for re-calibration. More detailde distribution comparison for each day can be found in \appendixname~\ref{appendix:cross-validation-d2d}.
\begin{figure}[htbp]
    \centering
    \begin{subfigure}{0.45\textwidth}
        \centering
        \includegraphics[width=0.9\linewidth]{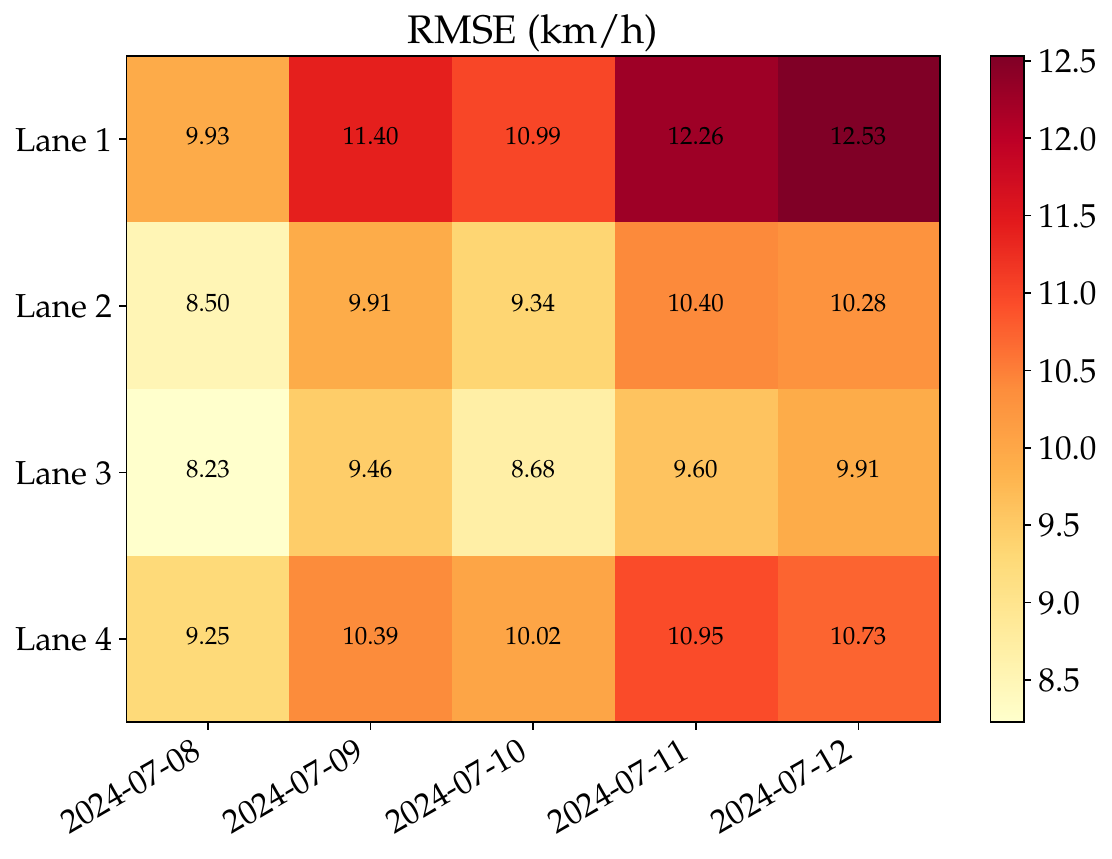}
        \caption{RMSE}
        \label{fig:cross-val-rmse}
    \end{subfigure}
    \begin{subfigure}{0.45\textwidth}
        \centering
        \includegraphics[width=0.9\linewidth]{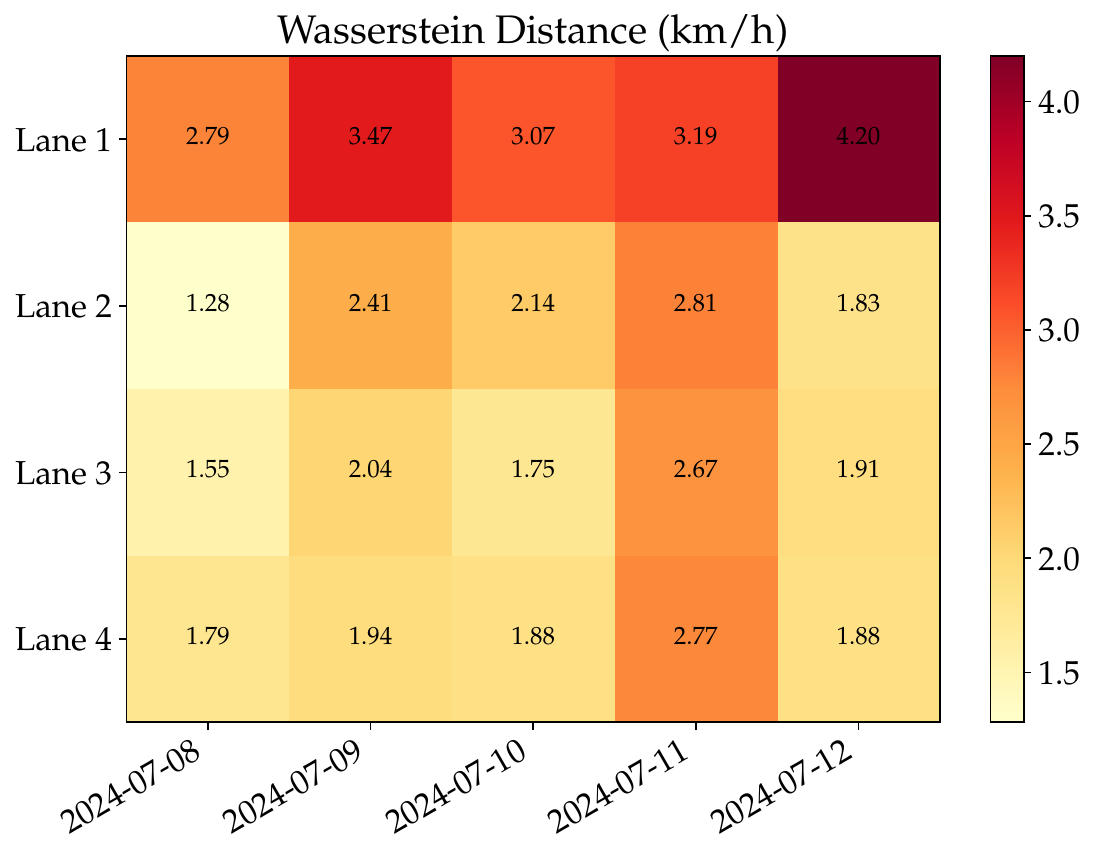}
        \caption{Wasserstein distance}
        \label{fig:cross-val-wasserstein}
    \end{subfigure}
    \caption{Cross-validation of calibrated parameters across multiple days. Each row represents the day used for calibration, and each column represents the day used for validation. (a) RMSE comparison heatmap. (b) Wasserstein distance comparison heatmap.}
    \label{fig:cross-validation-compared}
\end{figure}
\section{Discussions}
\label{sec:discussions}

\subsection{Convexity and reproducibility of the calibration problem}
Due to the existence of the nonconvex terms $\tanh$ and $\min$ in~\eqref{eq:nonconvex} and the Gaussian Kernel in~\eqref{eq:kernel1} and~\eqref{eq:kernel2}, the optimization problem is non-convex. As a result, it may converge to a local minimum if the initial conditions are not chosen carefully. As discussed in \cite{ahn2022reproducibility}, reproducibility has a connection with the convexity of the problem. Having different initial conditions might not give the same calibration results. In our case, we suggested the initial guess based on the parameters suggested by \cite{treiber2011reconstructing}. We also examined the convergence of the calibration across different lanes. The results indicate that, given the initial guess suggested by \cite{treiber2011reconstructing}, the calibration consistently converges. However, due to the non-convexity of the problem, the calibration results may vary if different initial conditions are used. This is a common issue in non-convex optimization problems, and it highlights the importance of carefully selecting initial conditions to ensure reproducibility.

\subsection{Day-to-day calibration}
We also calibrate the ASM parameters separately for each of the five days (July 8, 10, 11, 12, 2024) in addition to the original calibration on July 9, 2024. Each day of the time-space diagram is shown in \figurename~\ref{fig:d2d}.

\begin{figure}[htbp]
    \centering
    \includegraphics[width=\linewidth]{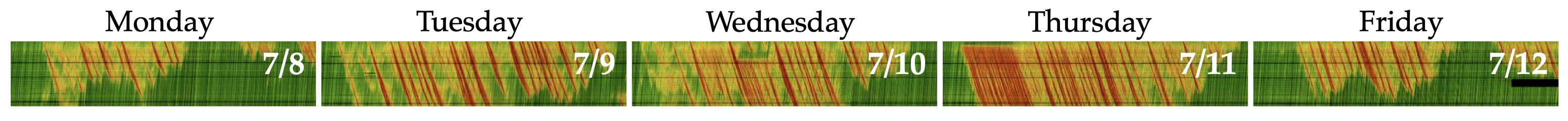}
    \caption{Multiple days collected for this study}
    \label{fig:d2d}
\end{figure}

The calibrated parameters for each day are shown in \figurename~\ref{fig:calibration_d2d}. Monday traffic congestion is mild, similar to Friday, the calibrated parameters are similar between these two days. Tuesday and Wednesday show more frequent congestion, the calibrated parameters are also similar between these two days. Thursday contains a lane-blocking crash event, and triggers more severe congestion, the calibrated parameters are different from other days. Overall, the calibrated parameters show some day-to-day variability, especially in response to traffic conditions. It is also observed that the calibrated parameters are quite different across lanes, especially for the wave propagation speeds $c_{cong}$ and $c_{free}$.

For practical applications, calibration performed on a representative day with typical traffic patterns can be reused for other days exhibiting similar patterns; in this study we use Tuesday (July 9, 2024). Meanwhile, lane-specific calibration is needed to capture the traffic patterns in each lane.

\begin{figure}[htbp]
    \centering
    \begin{subfigure}[b]{0.49\textwidth}
        \centering
        \includegraphics[width=\textwidth]{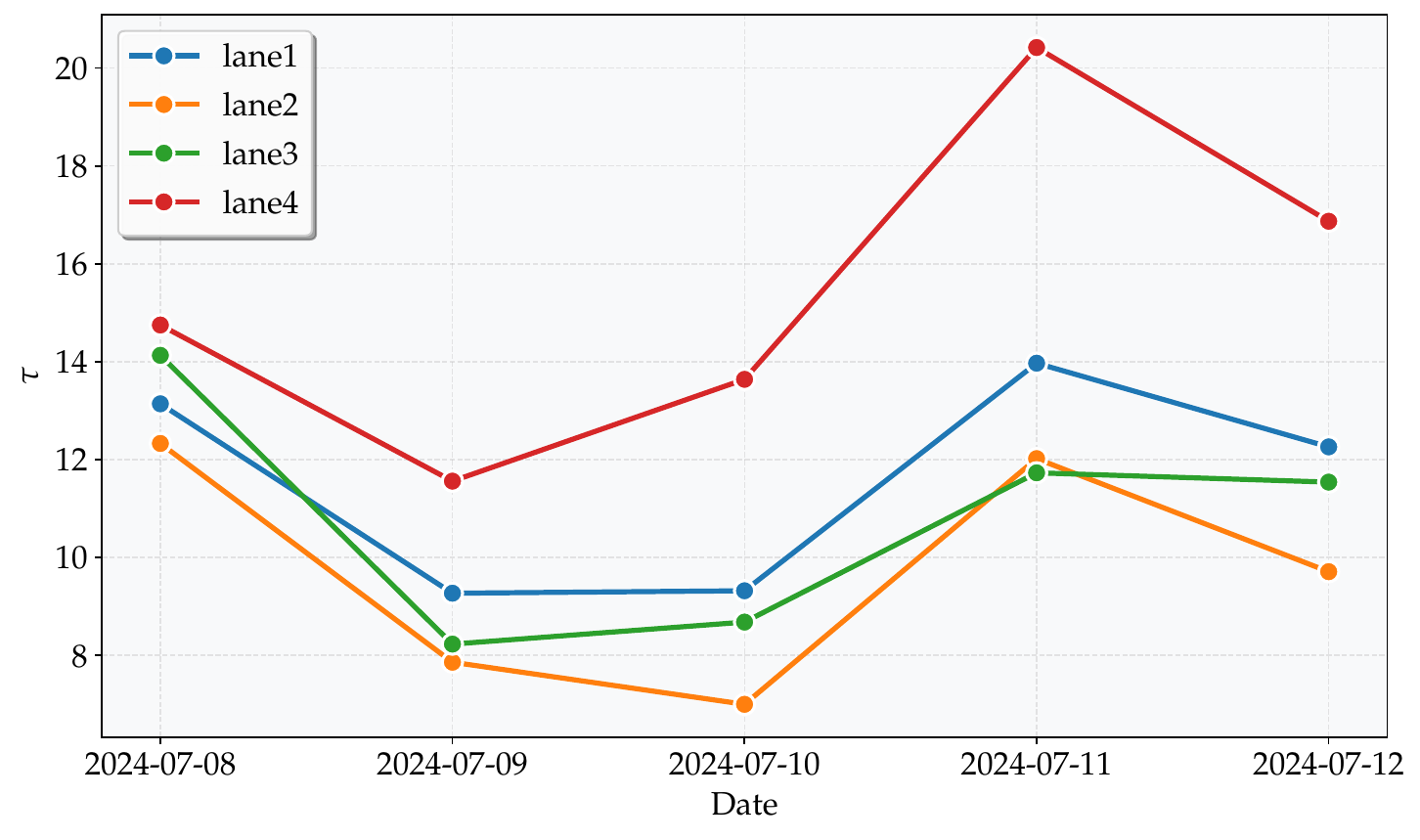}
        \caption{$\tau$ (seconds)}
        \label{fig:tau}
    \end{subfigure}
    \hfill
    \begin{subfigure}[b]{0.49\textwidth}
        \centering
        \includegraphics[width=\textwidth]{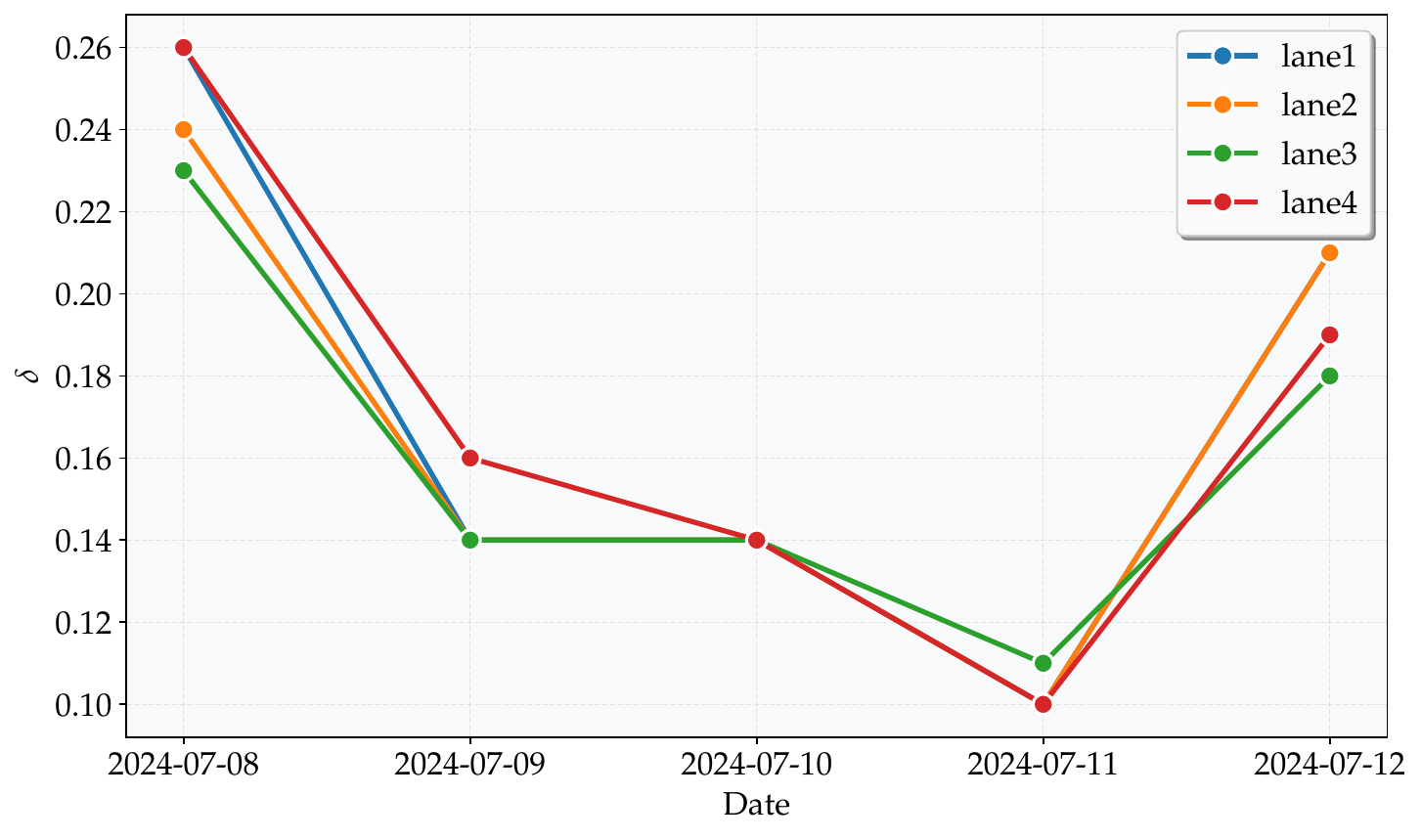}
        \caption{$\delta$ (km)}
        \label{fig:delta}
    \end{subfigure}
    
    \vspace{1em} %
    
    \begin{subfigure}[b]{0.49\textwidth}
        \centering
        \includegraphics[width=\textwidth]{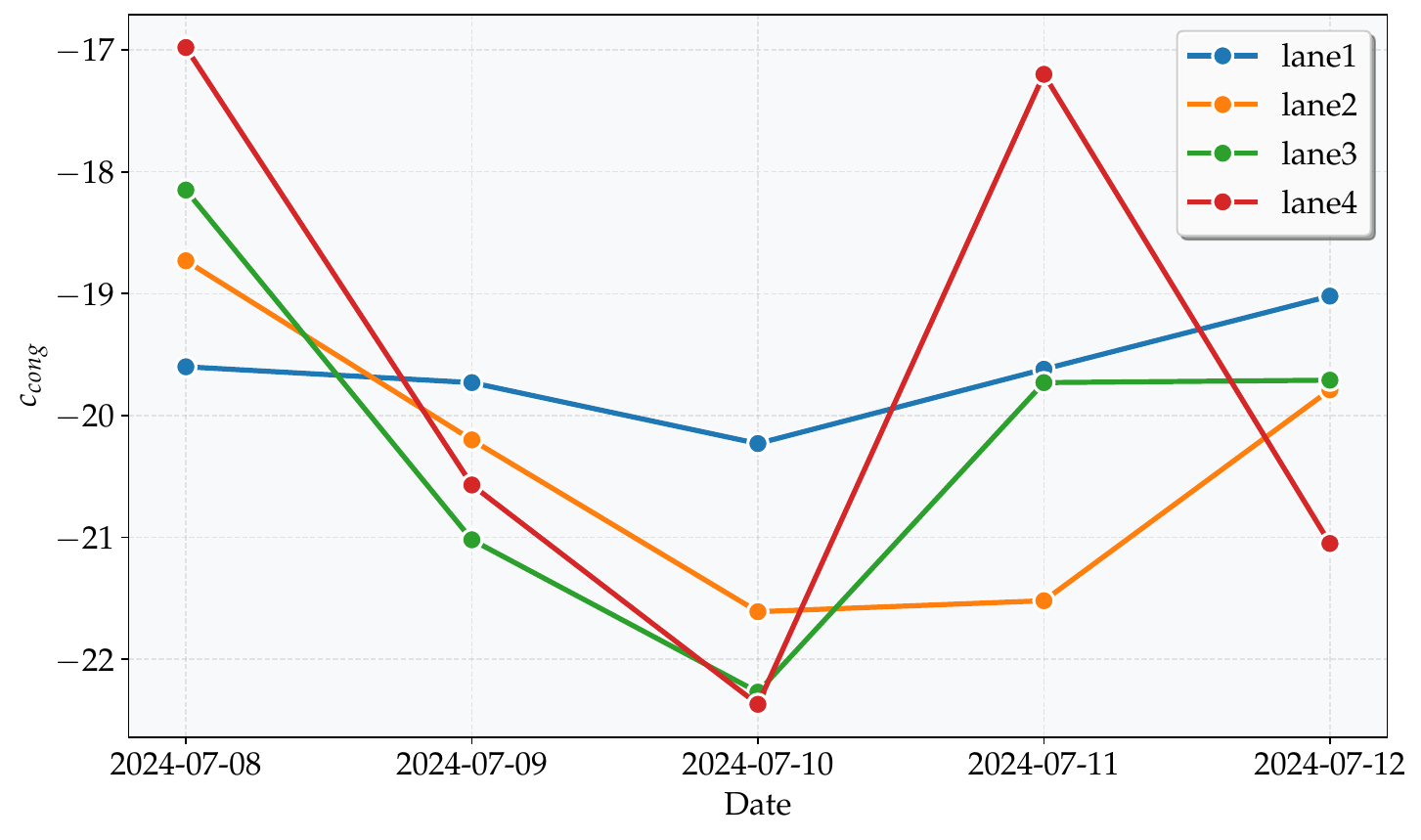}
        \caption{$c_\text{cong}$ (km/h)}
        \label{fig:c_cong}
    \end{subfigure}
    \hfill
    \begin{subfigure}[b]{0.49\textwidth}
        \centering
        \includegraphics[width=\textwidth]{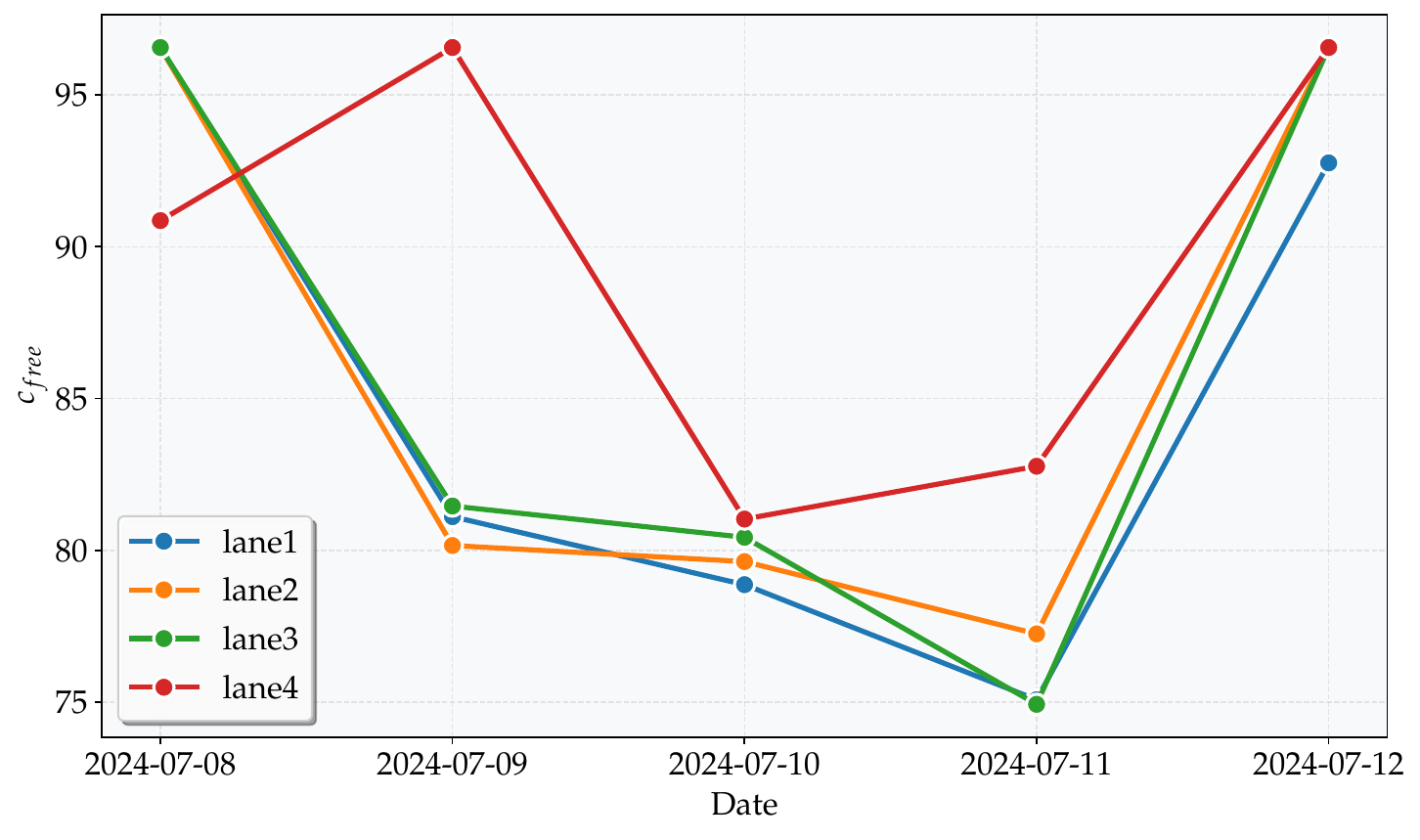}
        \caption{$c_\text{free}$ (km/h)}
        \label{fig:c_free}
    \end{subfigure}

    \vspace{1em} %

    \begin{subfigure}[b]{0.49\textwidth}
        \centering
        \includegraphics[width=\textwidth]{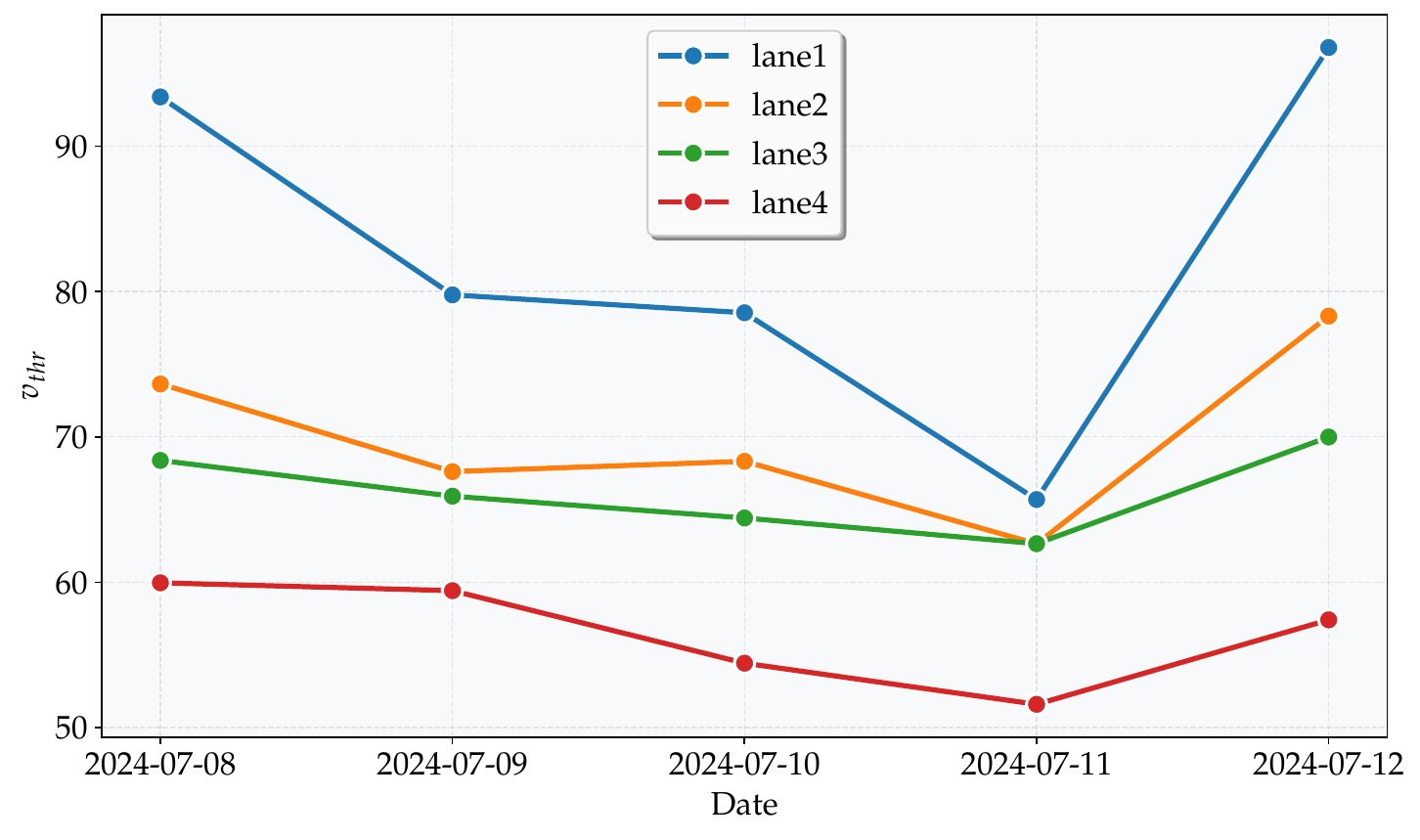}
        \caption{$V_\text{thr}$ (km/h)}
        \label{fig:v_thr}
    \end{subfigure}
    \hfill
    \begin{subfigure}[b]{0.49\textwidth}
        \centering
        \includegraphics[width=\textwidth]{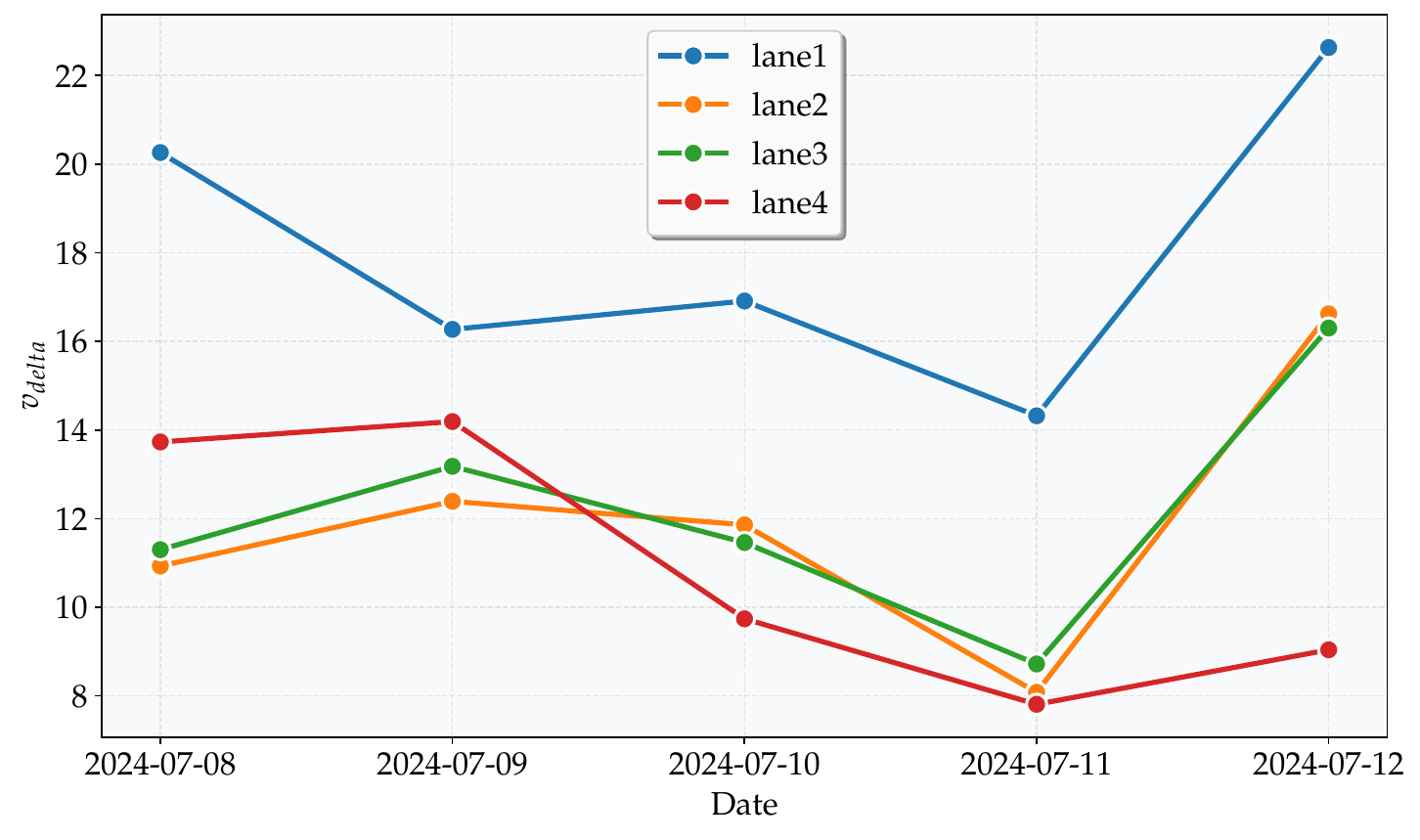}
        \caption{$\Delta V$ (km/h)}
        \label{fig:delta_v}
    \end{subfigure}
    
    \caption{Calibrated parameters over multiple days}
    \label{fig:calibration_d2d}
\end{figure}
\subsection{Use cases}
In this section, we demonstrate the use cases of the calibrated parameters and the implementation of ASM. While ground truth data is not available for these scenarios, the focus here is to illustrate the scalability and applicability of this open-source ASM implementation across different geographic locations and traffic conditions. Note that the traffic state reconstructed in this section is not validated with ground truth. The primary goal is to highlight how the provided code can be efficiently reused for large-scale freeway sensor networks with rapid deployment.
\subsubsection{Application to I-24 SMART Corridor}
I-24 SMART Corridor \cite{coursey2024ft,zhang2025real} is a 27.36 km corridor between Nashville and Murfreesboro, Tennessee. The corridor is equipped with the Wavetronix HD sensors, which provide the 30-second aggregated speed, occupancy, and volume data. The sensors are spaced at an average interval of 483 meters. The calibration data is collected from I-24 MOTION, which is a 6.76 km testbed located along the same interstate roadway. 

\begin{figure}[htbp]
    \centering
    \begin{minipage}{0.80\linewidth}
        \centering
        \includegraphics[width=\linewidth]{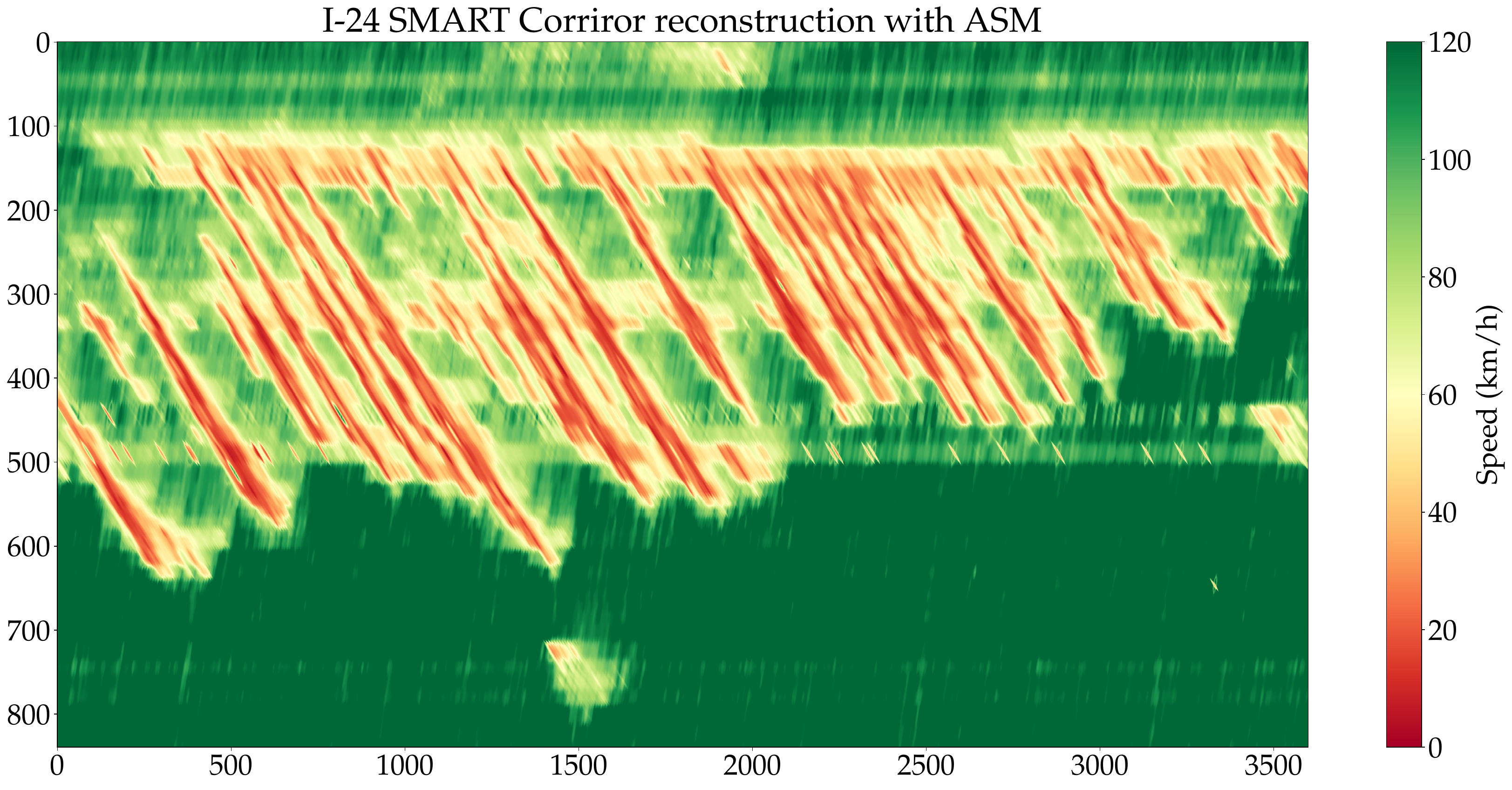}
        \subcaption{ASM reconstruction with calibrated parameters}
    \end{minipage}
    \\[1em]
    \begin{minipage}{0.80\linewidth}
        \centering
        \includegraphics[width=\linewidth]{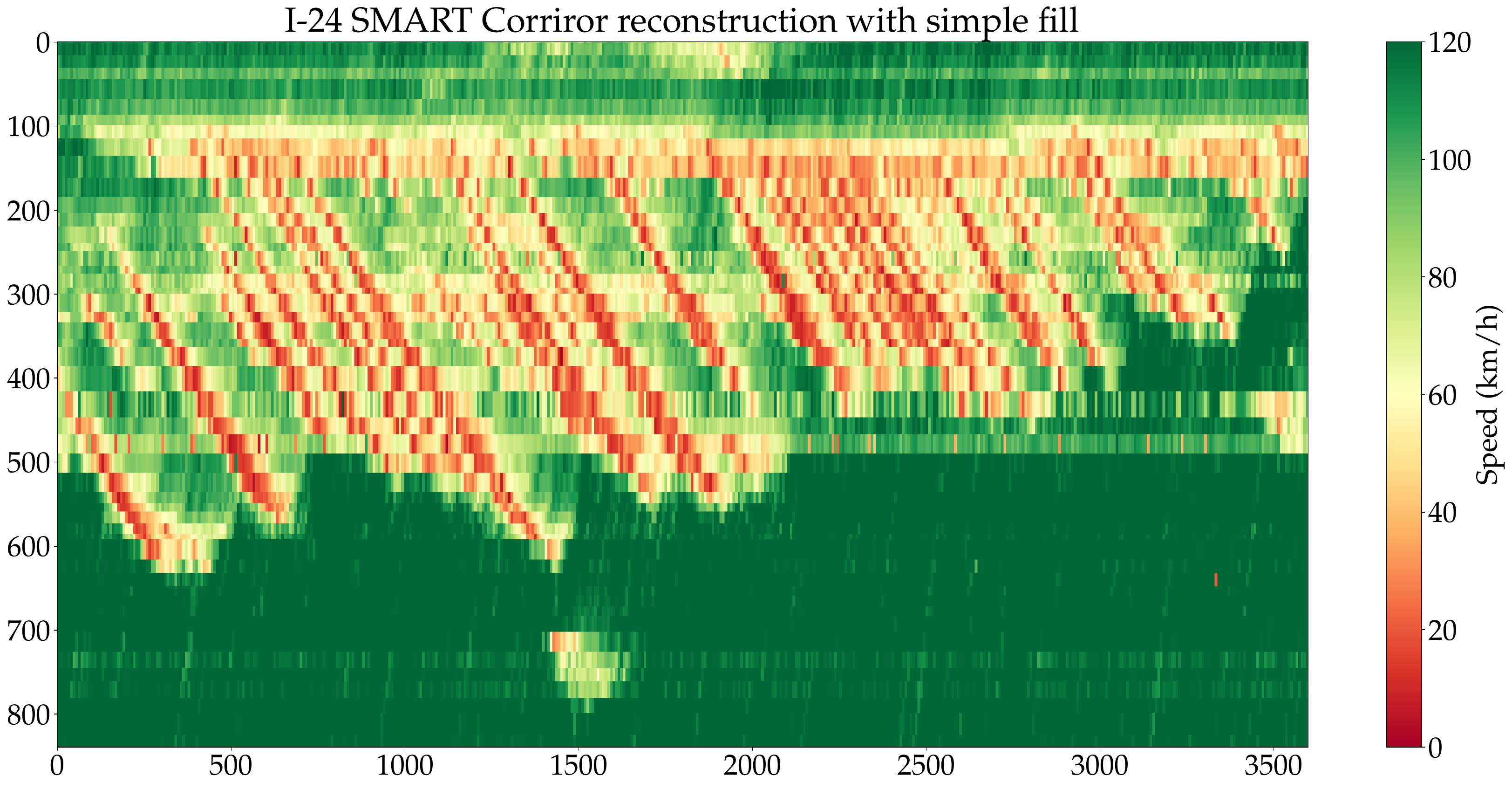}
        \subcaption{Simple reconstruction using last observed speed}
        \label{fig:i24simple_reconstruction}
    \end{minipage}
    \\[1em]
    \caption{I-24 SMART Corridor speed reconstruction for lane 1 using calibrated ASM parameters. Data covers 27.36 km on July 9th, 2024, from 6:00 AM to 10:00 AM. CPU processing time: 1.07 seconds.}
    \label{fig:i24}
\end{figure}

Figure~\ref{fig:i24} presents the speed reconstruction for lane 1 of the I-24 SMART Corridor using the calibrated ASM parameters. In comparison to the simple reconstruction method shown in Figure~\ref{fig:i24simple_reconstruction}, the ASM-based approach yields a more accurate and detailed depiction of traffic conditions. Notably, spatio-temporal wave patterns are more clearly delineated, enabling improved identification of congestion and slowdowns for operational decision-making. The computational time for this reconstruction task (approximately 27.36 kilometers and 4 hours) is approximately 1.07 seconds in CPU time, demonstrating the efficiency of the ASM implementation for real-time applications.

\subsubsection{Application to Virginia Corridor}
The Virginia Department of Transportation operates a variable speed limit (VSL) corridor on I-95 northbound which spans 24.14 km (mile markers 115 to 130) and has been operational since June 2022. This section of roadway sits between Richmond, VA and Washington, DC near the city of Fredericksburg, VA. The corridor consists of 24 detection sites that collect per-vehicle record (PVR) data, which provide the entry timestamp, lane, length, speed, and duration measures of the vehicles which pass by the sensor; this allows for critical speed, occupancy, and volume metrics to be sent in 30 second intervals to the operating algorithm that makes posted speed limit recommendations. The sensor network is made up of Wavetronix SmartSensor HD radar detectors, which are the same type of sensors used in the I-24 SMART Corridor.
The system then posts speed limit updates every minute through 48 roadside digital signs, strategically placed with an average spacing of 0.97 km. We use the data collected from April 9th, 2025, from 11:05 AM to 3:35 PM for the reconstruction task. Lane aggregated speed data is used for the reconstruction.

In this case, the reconstruction task is to reconstruct the speed data for I-95 northbound, with approximately 24.14 kilometers of roadway and 4.5 hours of data. The reconstruction is performed using the same parameters as for I-24 SMART Corridor.

\begin{figure}[htbp]
    \centering
    \begin{minipage}{0.90\linewidth}
        \centering
        \includegraphics[width=\linewidth]{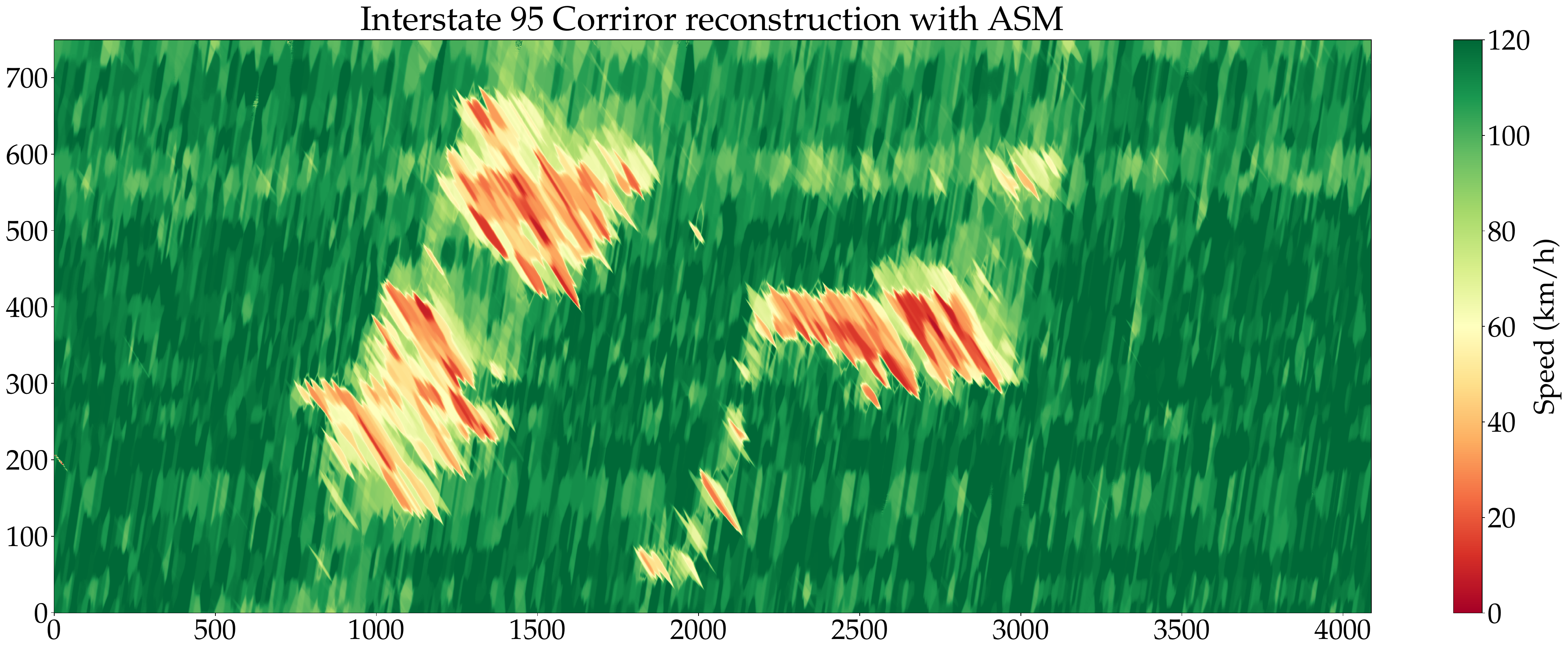}
        \subcaption{ASM reconstruction with calibrated parameters}
    \end{minipage}
    \\
    \begin{minipage}{0.90\linewidth}
        \centering
        \includegraphics[width=\linewidth]{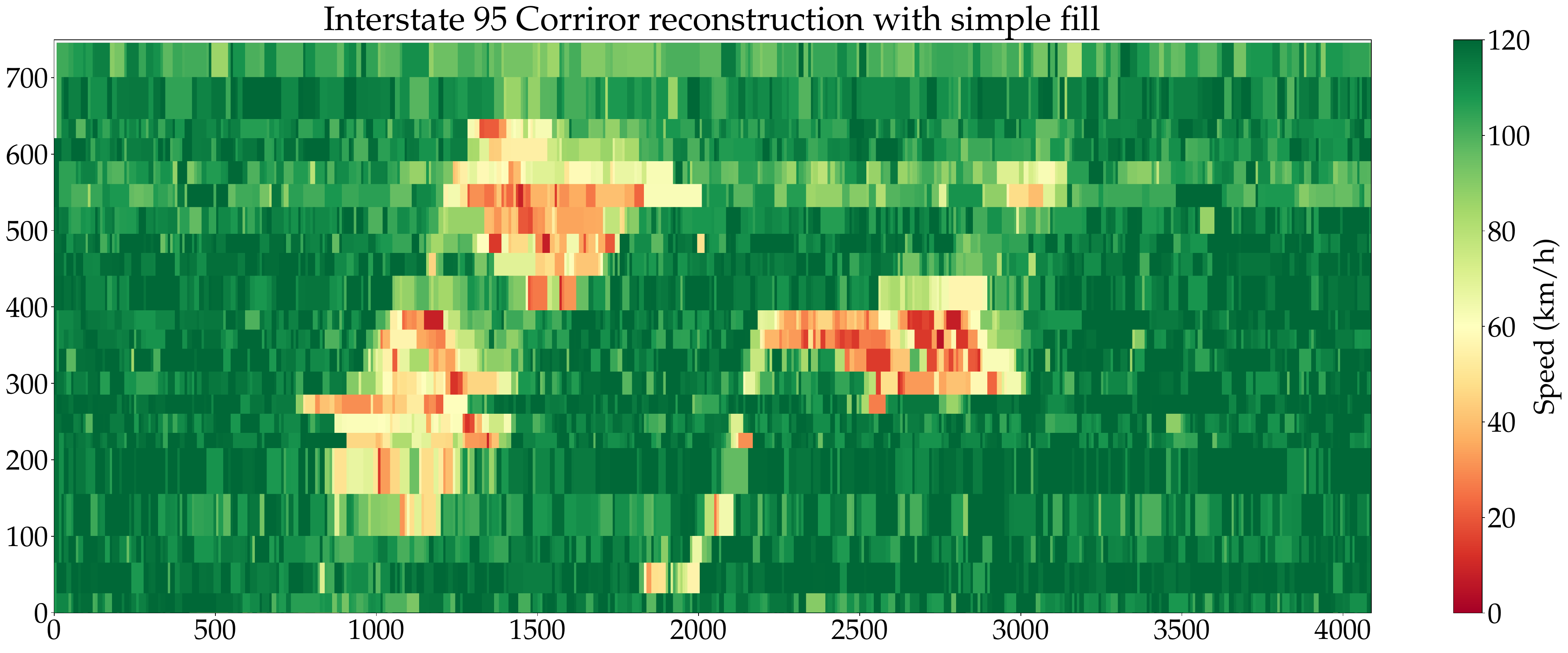}
        \subcaption{Simple reconstruction using last observed speed}
    \end{minipage}
    \caption{Speed data reconstruction at Interstate 95 Northbound with parameters set to Lane 1 for Mile Marker 115 to Mile Marker 130, covering 24.1 km, with CPU processing time of 1.30 seconds.}
    \label{fig:i95}
\end{figure}

As shown in Figure~\ref{fig:i95}, the data quality on the Virginia corridor is generally lower than that of the I-24 SMART Corridor, with more frequent and irregularly distributed missing data. Despite these challenges, the ASM implementation remains effective, providing reasonable reconstructions even in the presence of substantial data gaps.
It is important to note that the reconstruction in this section is purely served as the illustration of the use case of the ASM implementation, and the parameters used here are not calibrated specifically for this corridor. We include this example mainly as a demonstration, and rigorous evaluation would require ground truth data, which is not available for this corridor at present. Floating vehicle data or drone-based local trajectory data could be used in future work to validate the reconstruction quality.

\subsection{Future improvements and extensions to ASM}
More recent study suggest the propagation speed may vary with traffic conditions \cite{coifman2023lwr, ji2024scalable}, this can be an interesting direction for future work to explore adaptive kernels that change based on local traffic states, but the trade-off between model complexity and computational efficiency should be carefully considered. 

Other extension could incorporate a finite set of more than two anisotropic kernels \cite{yang2022generalized} to better adapt to traffic states and capture complex traffic dynamics.

ASM currently assumes a stationary kernel (constant in time and space), which simplifies estimation and reduces the number of parameters. However, traffic dynamics often vary across locations (e.g., sensor spacing, road geometry) and over time, so relaxing the stationary assumption by using non-stationary or locally adaptive kernels, could yield more accurate representations of real-world traffic. Future work could investigate these extensions while explicitly considering the trade-offs between modeling flexibility, interpretability, and computational efficiency.
\section{Conclusion}
\label{sec:conclusion}
In this article, we present an open-source implementation of the Adaptive Smoothing Method (ASM) for freeway speed reconstruction, which is a kernel-based method that reconstructs the speed data from sparse sensor observations. The implementation is designed to be efficient and scalable, making it suitable for real-time freeway operations. We calibrate the method in the context of the I-24 MOTION testbed, which provides the ground truth mean speed field data for calibration. 

We would also like to highlight how open data and code accelerate research progress. The source code released with this article is the result of several iterations to improve the ASM implementation \cite{ji2024virtual}. Earlier versions were either incorrect or not efficient enough for large-scale freeway networks. The current implementation benefits from early community feedback on our preliminary releases, which helps identify errors and improve computational efficiency. We hope that this open-source implementation will not only facilitate further research and development in freeway speed reconstruction and related applications, but also pass the momentum of open-source development to the community to continue building upon existing work.

\appendix
\section{Nonlinear adaptive speed filter comparison}
\label{appendix:filter}
The nonlinear adaptive speed filter is designed to smoothly transition between free-flow and congested traffic regimes with an s-shaped nonlinear function. ASM originally used a $tanh$ function for this purpose. Other s-shaped functions, such as the sigmoid function, could also be used. We have added a side-by-side comparison of the two functions in Figure~\ref{fig:side_by_side} below to illustrate their similarities and differences. 
    
    With the $tanh$-based filter as:
    \begin{equation}
    w(v) = \frac{1}{2}\left[1 + \tanh\left(\frac{v_c - v}{\Delta v}\right)\right],
    \end{equation}
    and the sigmoid-based filter as:
    \begin{equation}
    w(v) = \frac{1}{1 + \exp\left(\frac{v - v_c}{\Delta v}\right)}.
    \end{equation}
    To make a fair comparison, both functions are centered at the same critical speed $v_c = 50$ km/h and have the same transition width $\Delta v = 10$ km/h.

\begin{figure}[htbp]
     \centering
     \begin{subfigure}[b]{0.48\textwidth}
         \centering
         \includegraphics[width=\linewidth]{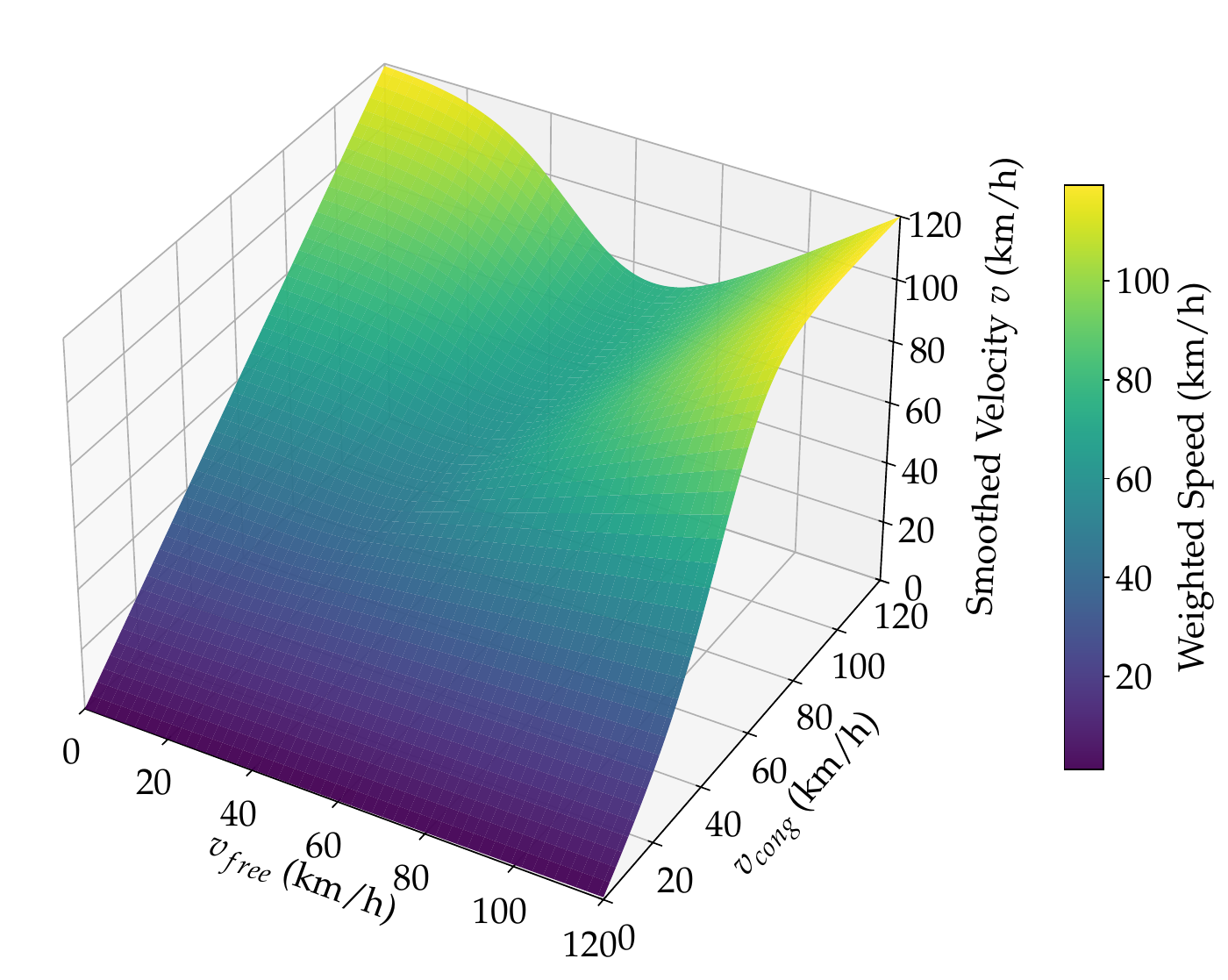}
         \caption{$\tanh$-based adaptive speed filter used in ASM}
     \end{subfigure}
     \hfill %
     \begin{subfigure}[b]{0.48\textwidth}
         \centering
         \includegraphics[width=\linewidth]{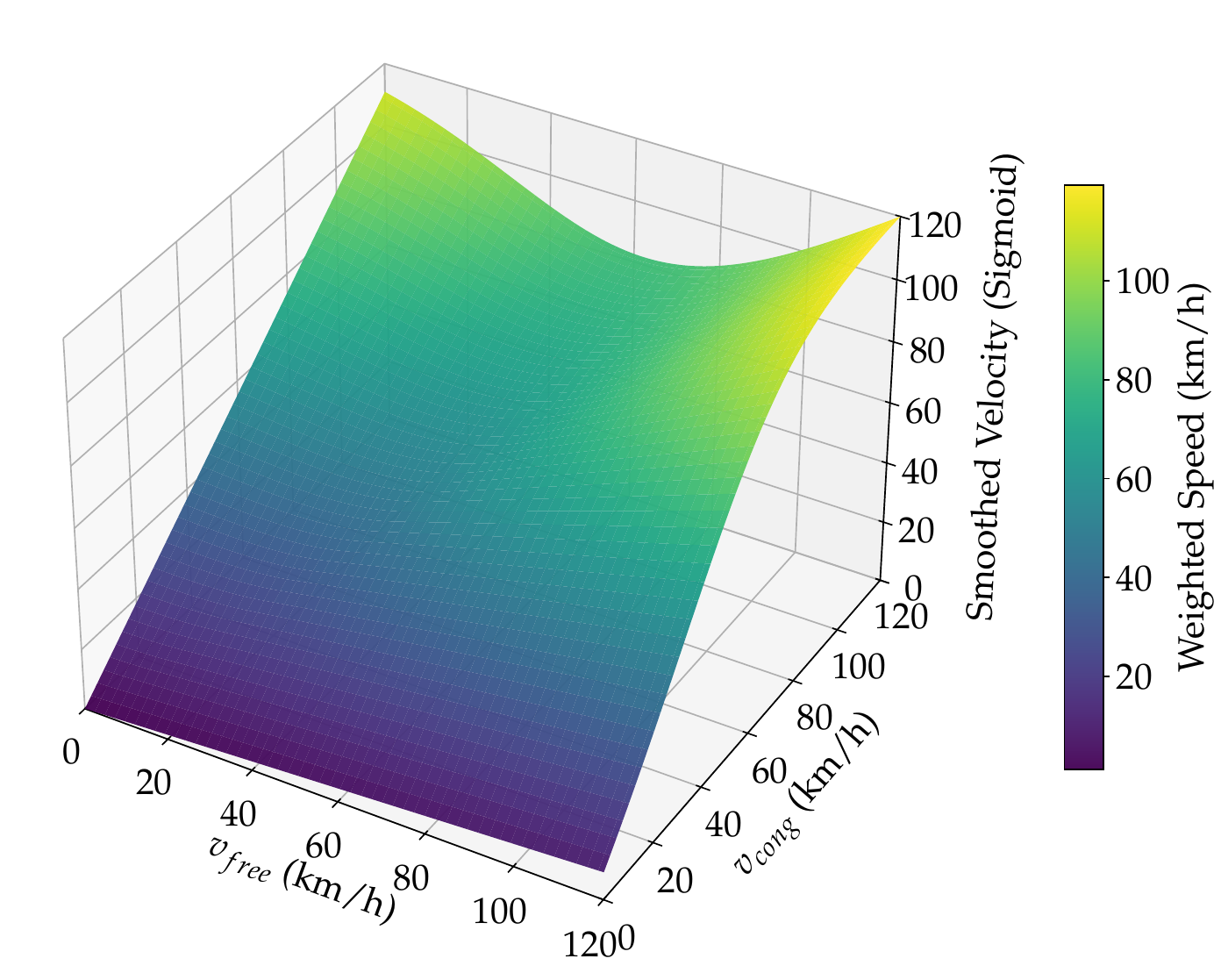}
         \caption{Sigmoid-based adaptive speed filter as an alternative}
     \end{subfigure}
     \caption{Comparison of adaptive speed filters used in ASM}
     \label{fig:side_by_side}
\end{figure}

We further show the difference between the mixed speed difference using the two s-shape filters in Figure~\ref{fig:comparison_tanh_sigmoid} below. $sigmoid$-based filter results in a slightly smoother transition compared to the $tanh$-based function. As can be seen from the difference heatmap, the differences in congested or free-flow regions are minimal, however, the differences are more pronounced near the critical speed boundary where the transition occurs. We add this comparison to Appendix in the manuscript.
\begin{figure}[htbp]
    \centering
    \includegraphics[width=0.8\linewidth]{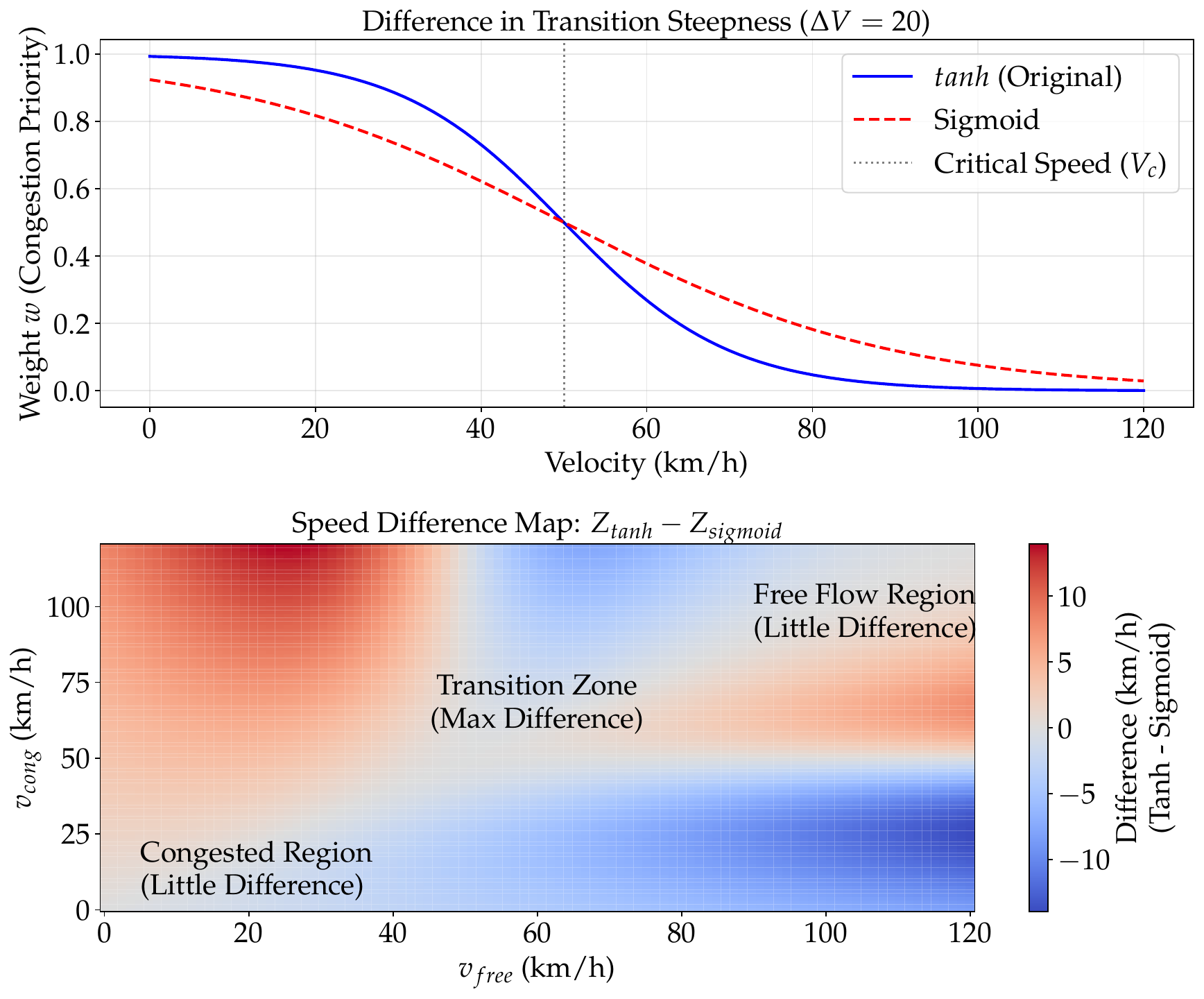}
    \caption{Comparison of mixed speed difference using $\tanh$-based and sigmoid-based filters}
    \label{fig:comparison_tanh_sigmoid}
\end{figure}

\section{Cross-validation for day-to-day speed distribution comparison}
\label{appendix:cross-validation-d2d}
We include more days for cross-validation of the July 9th, 2024 calibrated parameters on the other days (July 8, 10, 11, 12, 2024). The results are shown in \figurename~\ref{fig:speed_histograms_lane1_2} and \figurename~\ref{fig:speed_histograms_lane3_4} below. The persistent low-speed region mismatch is mainly caused by sensors reporting the time-mean speed, which overestimates the true space-mean speed. ASM itself is unable to correct this bias.
\begin{figure}[htbp]
    \centering
    
    \begin{subfigure}[b]{0.48\textwidth}
        \centering
        \includegraphics[width=\linewidth]{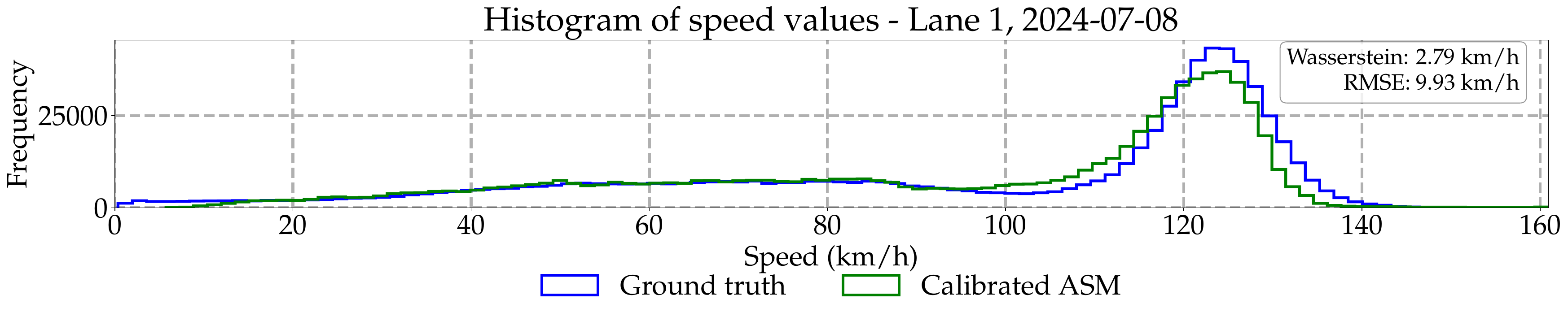}
        \caption{Lane 1 -- July 08}
        \label{fig:l1-08}
    \end{subfigure}
    \hfill
    \begin{subfigure}[b]{0.48\textwidth}
        \centering
        \includegraphics[width=\linewidth]{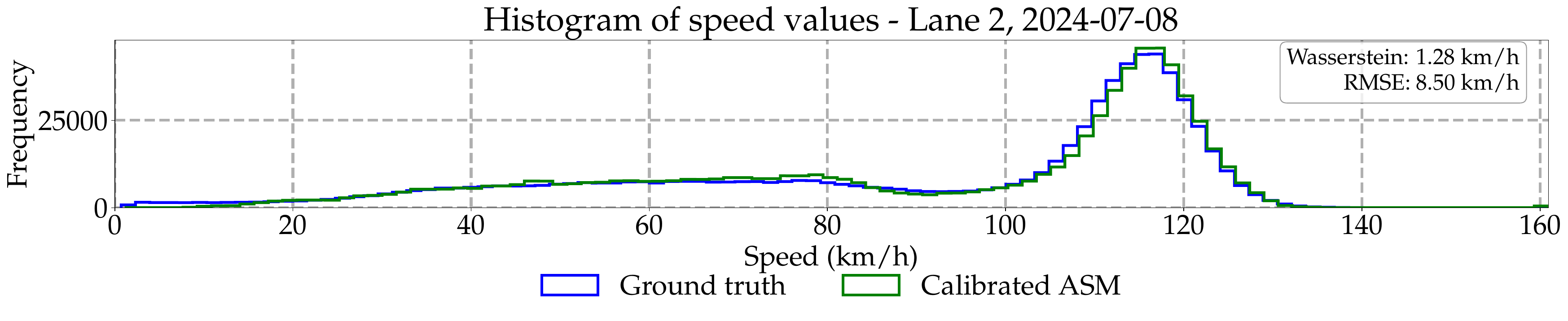}
        \caption{Lane 2 -- July 08}
        \label{fig:l2-08}
    \end{subfigure}
    
    \vspace{0cm} %
    
    \begin{subfigure}[b]{0.48\textwidth}
        \centering
        \includegraphics[width=\linewidth]{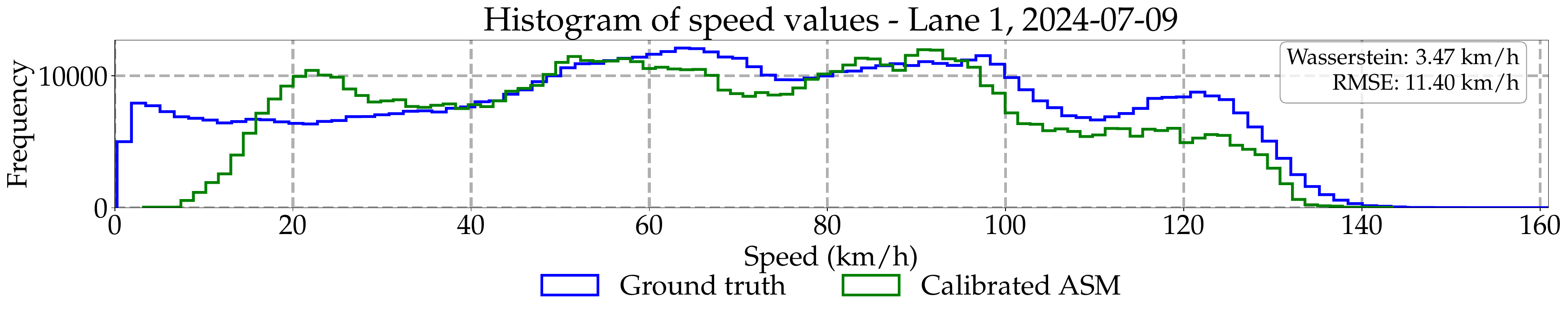}
        \caption{Lane 1 -- July 09}
        \label{fig:l1-09}
    \end{subfigure}
    \hfill
    \begin{subfigure}[b]{0.48\textwidth}
        \centering
        \includegraphics[width=\linewidth]{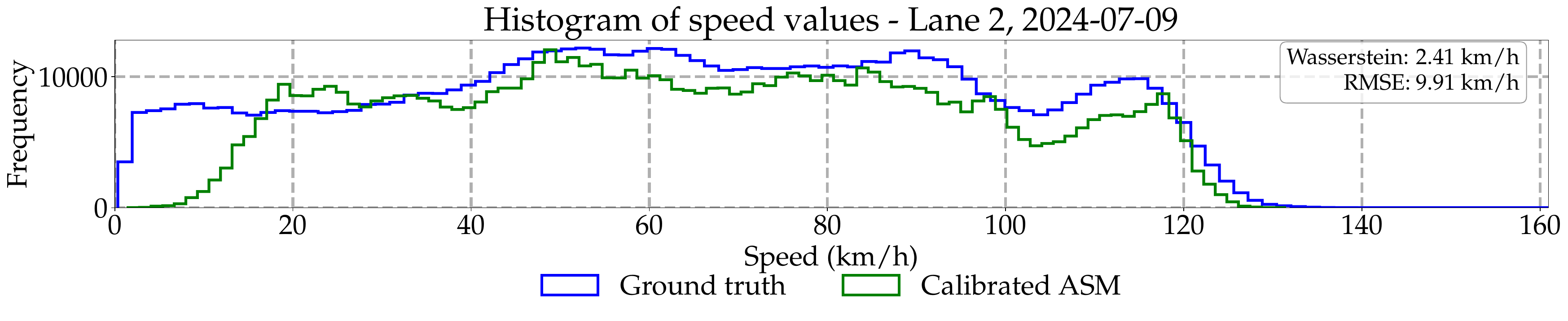}
        \caption{Lane 2 -- July 09}
        \label{fig:l2-09}
    \end{subfigure}
    
    \vspace{0cm}
    
    \begin{subfigure}[b]{0.48\textwidth}
        \centering
        \includegraphics[width=\linewidth]{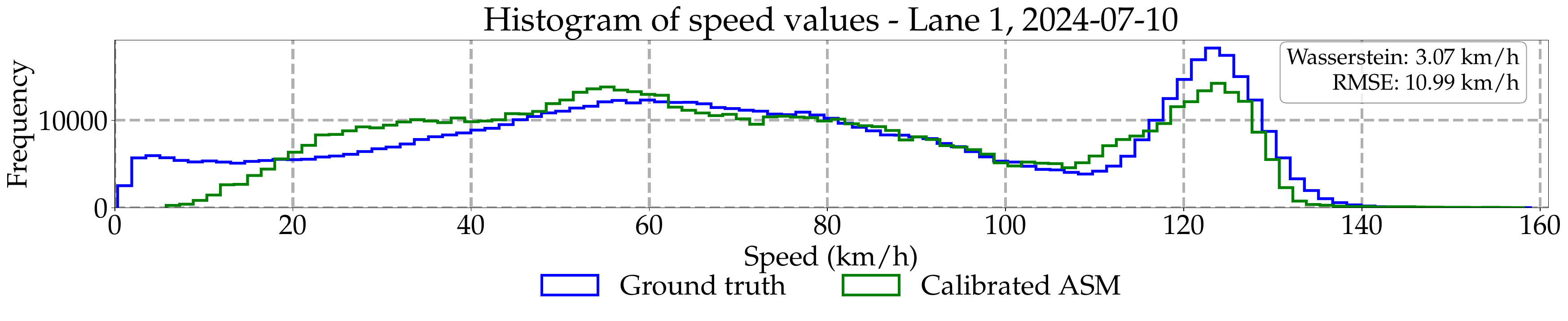}
        \caption{Lane 1 -- July 10}
        \label{fig:l1-10}
    \end{subfigure}
    \hfill
    \begin{subfigure}[b]{0.48\textwidth}
        \centering
        \includegraphics[width=\linewidth]{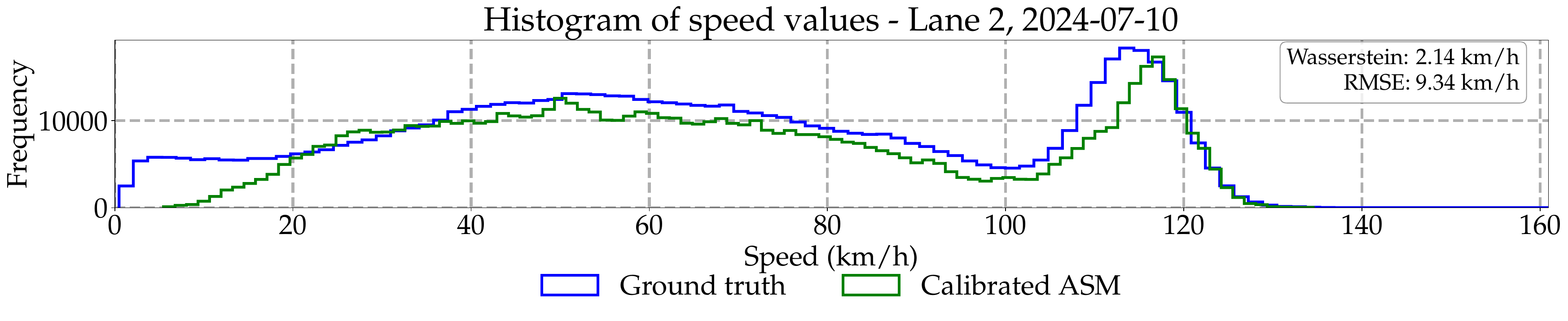}
        \caption{Lane 2 -- July 10}
        \label{fig:l2-10}
    \end{subfigure}

    \vspace{0cm}

    \begin{subfigure}[b]{0.48\textwidth}
        \centering
        \includegraphics[width=\linewidth]{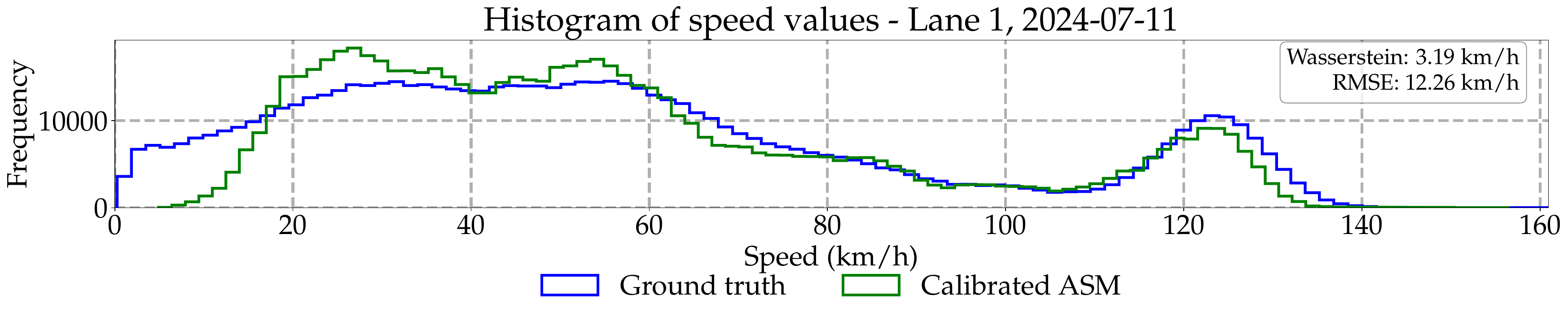}
        \caption{Lane 1 -- July 11}
        \label{fig:l1-11}
    \end{subfigure}
    \hfill
    \begin{subfigure}[b]{0.48\textwidth}
        \centering
        \includegraphics[width=\linewidth]{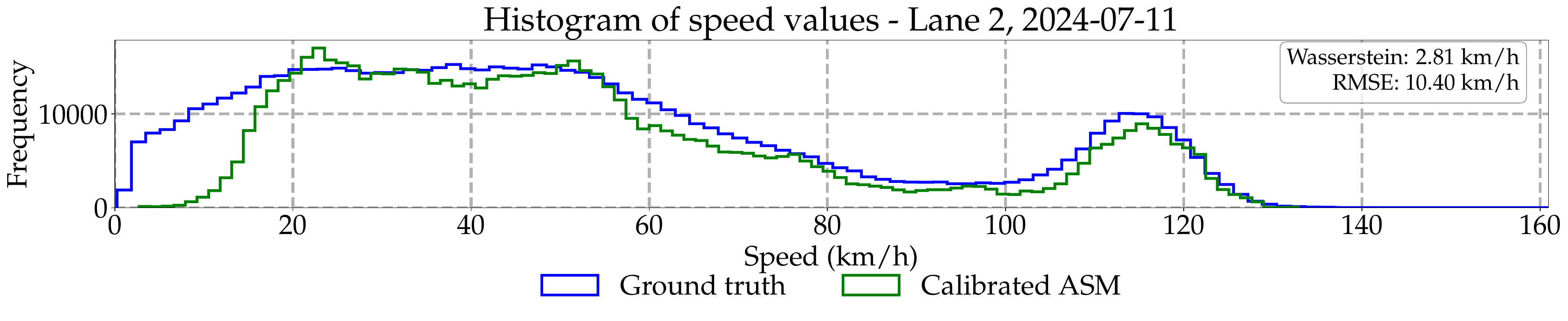}
        \caption{Lane 2 -- July 11}
        \label{fig:l2-11}
    \end{subfigure}

    \vspace{0cm}

    \begin{subfigure}[b]{0.48\textwidth}
        \centering
        \includegraphics[width=\linewidth]{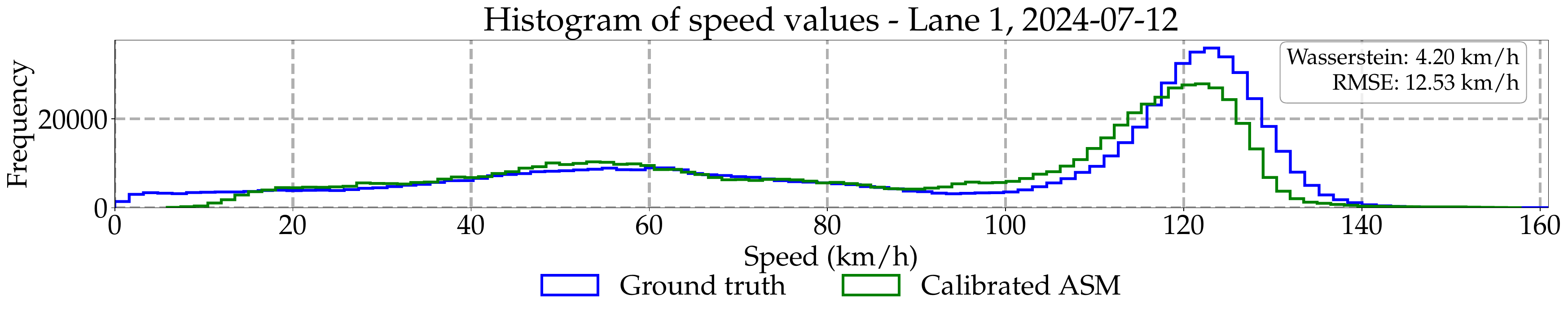}
        \caption{Lane 1 -- July 12}
        \label{fig:l1-12}
    \end{subfigure}
    \hfill
    \begin{subfigure}[b]{0.48\textwidth}
        \centering
        \includegraphics[width=\linewidth]{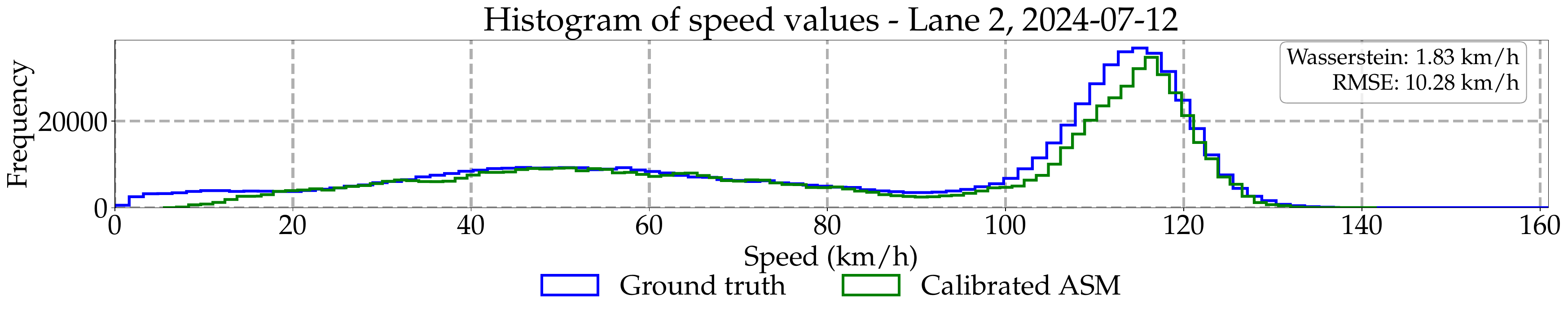}
        \caption{Lane 2 -- July 12}
        \label{fig:l2-12}
    \end{subfigure}
    
    \caption{Comparison of Speed Value Histograms. The left column shows Lane 1 and the right column shows Lane 2, spanning from July 08 to July 12, 2024. Ground truth is shown in blue and Calibrated ASM in green.}
    \label{fig:speed_histograms_lane1_2}
\end{figure}

\begin{figure}[htbp]
    \centering
    
    \begin{subfigure}[b]{0.48\textwidth}
        \centering
        \includegraphics[width=\linewidth]{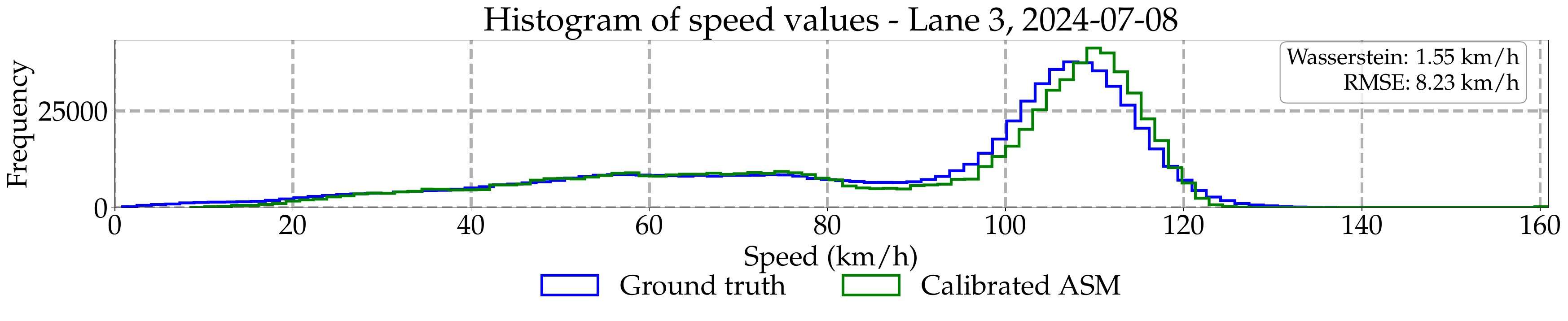}
        \caption{Lane 3 -- July 08}
        \label{fig:l3-08}
    \end{subfigure}
    \hfill
    \begin{subfigure}[b]{0.48\textwidth}
        \centering
        \includegraphics[width=\linewidth]{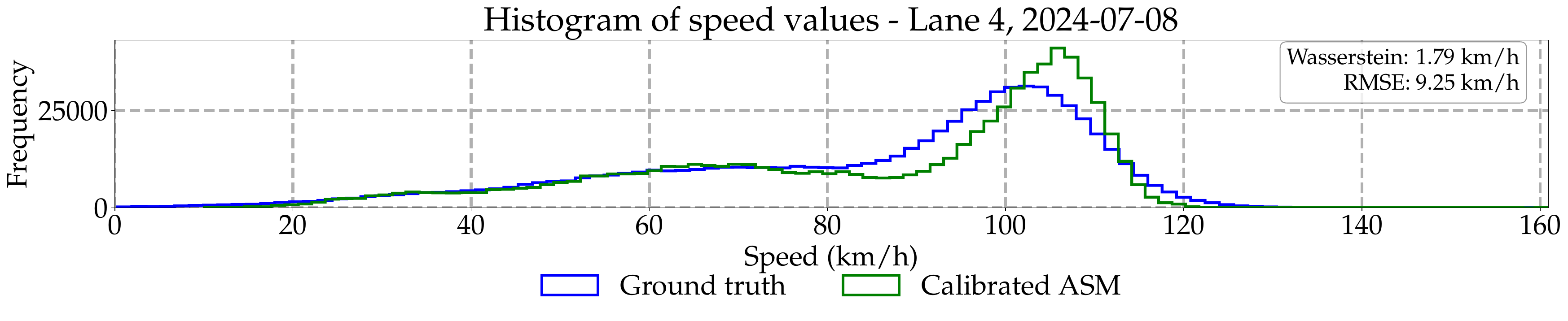}
        \caption{Lane 4 -- July 08}
        \label{fig:l4-08}
    \end{subfigure}
    
    \vspace{0cm} %
    
    \begin{subfigure}[b]{0.48\textwidth}
        \centering
        \includegraphics[width=\linewidth]{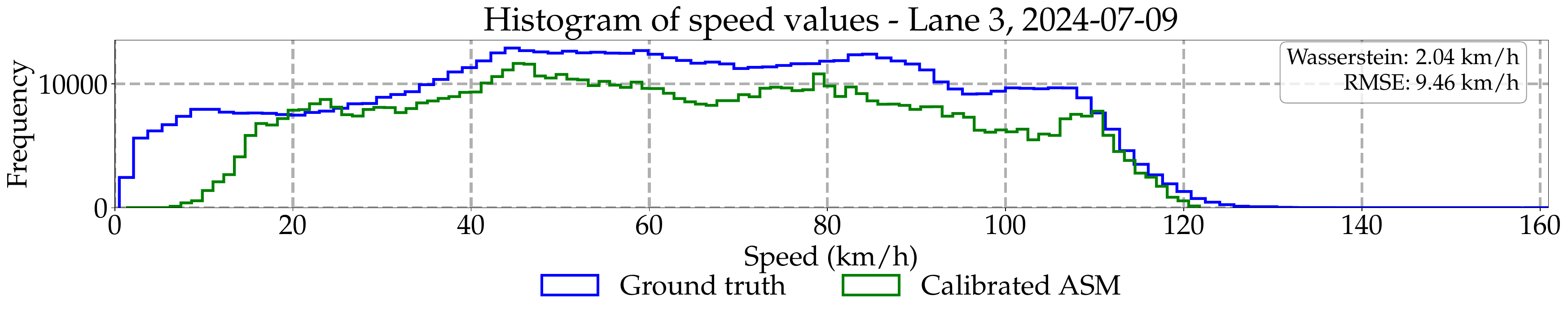}
        \caption{Lane 3 -- July 09}
        \label{fig:l3-09}
    \end{subfigure}
    \hfill
    \begin{subfigure}[b]{0.48\textwidth}
        \centering
        \includegraphics[width=\linewidth]{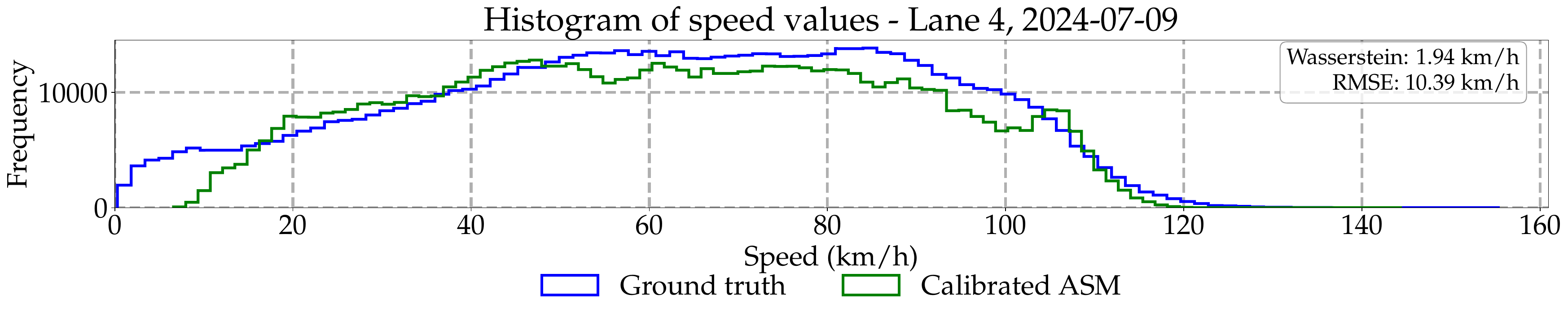}
        \caption{Lane 4 -- July 09}
        \label{fig:l4-09}
    \end{subfigure}
    
    \vspace{0cm}
    
    \begin{subfigure}[b]{0.48\textwidth}
        \centering
        \includegraphics[width=\linewidth]{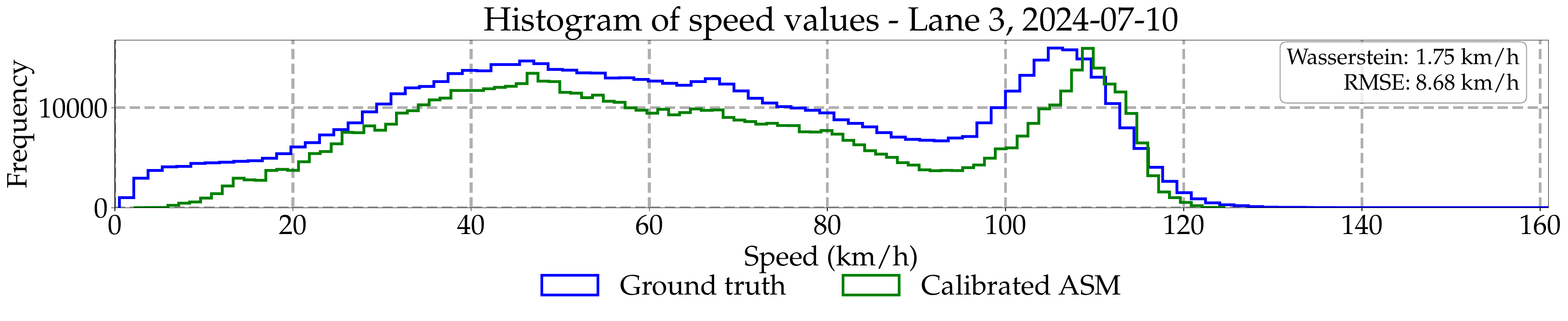}
        \caption{Lane 3 -- July 10}
        \label{fig:l3-10}
    \end{subfigure}
    \hfill
    \begin{subfigure}[b]{0.48\textwidth}
        \centering
        \includegraphics[width=\linewidth]{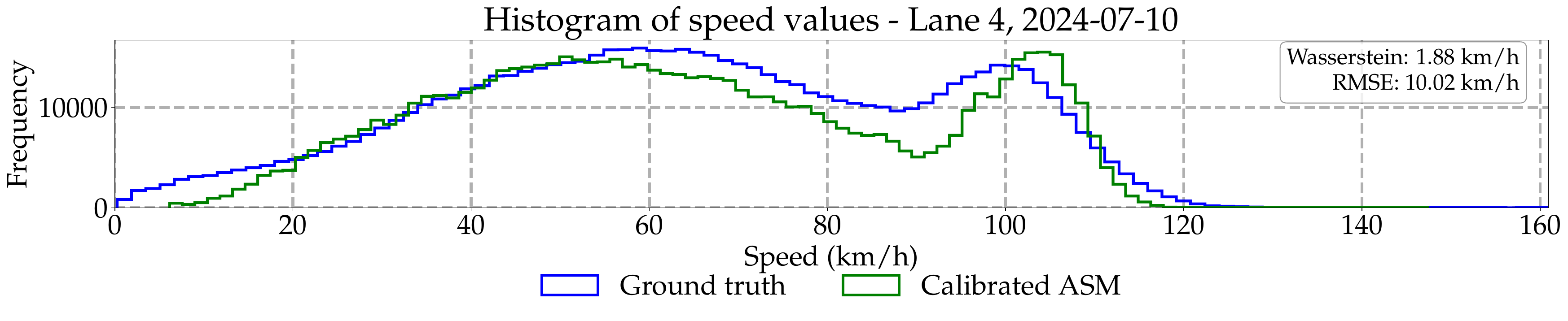}
        \caption{Lane 4 -- July 10}
        \label{fig:l4-10}
    \end{subfigure}

    \vspace{0cm}

    \begin{subfigure}[b]{0.48\textwidth}
        \centering
        \includegraphics[width=\linewidth]{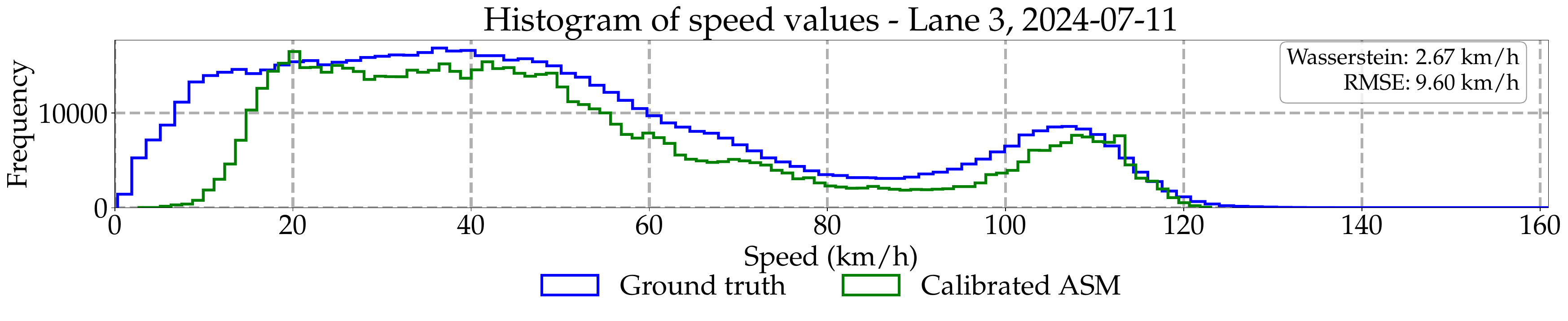}
        \caption{Lane 3 -- July 11}
        \label{fig:l3-11}
    \end{subfigure}
    \hfill
    \begin{subfigure}[b]{0.48\textwidth}
        \centering
        \includegraphics[width=\linewidth]{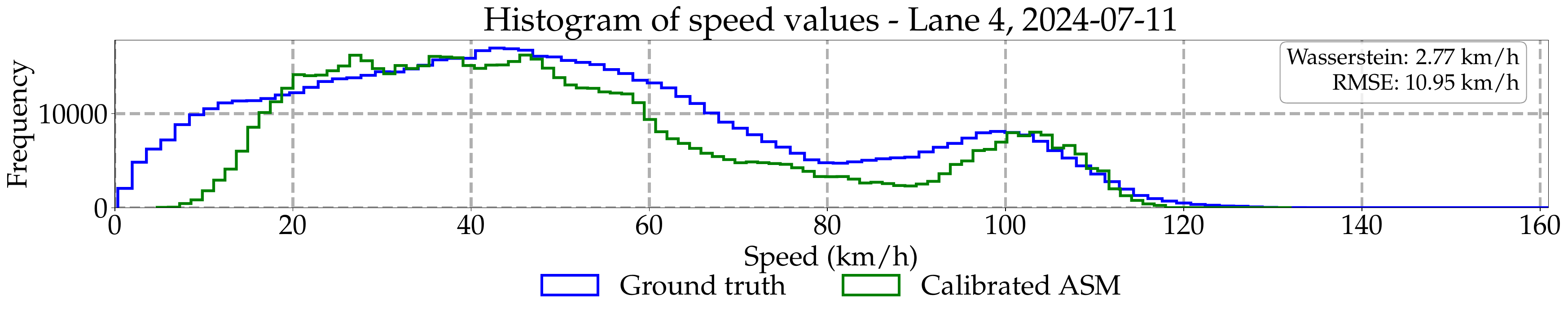}
        \caption{Lane 4 -- July 11}
        \label{fig:l4-11}
    \end{subfigure}

    \vspace{0cm}

    \begin{subfigure}[b]{0.48\textwidth}
        \centering
        \includegraphics[width=\linewidth]{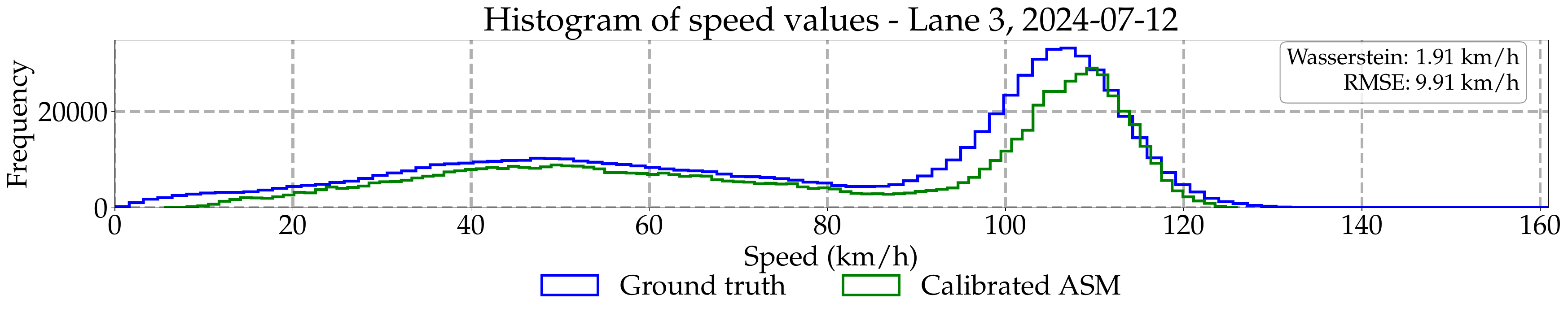}
        \caption{Lane 3 -- July 12}
        \label{fig:l3-12}
    \end{subfigure}
    \hfill
    \begin{subfigure}[b]{0.48\textwidth}
        \centering
        \includegraphics[width=\linewidth]{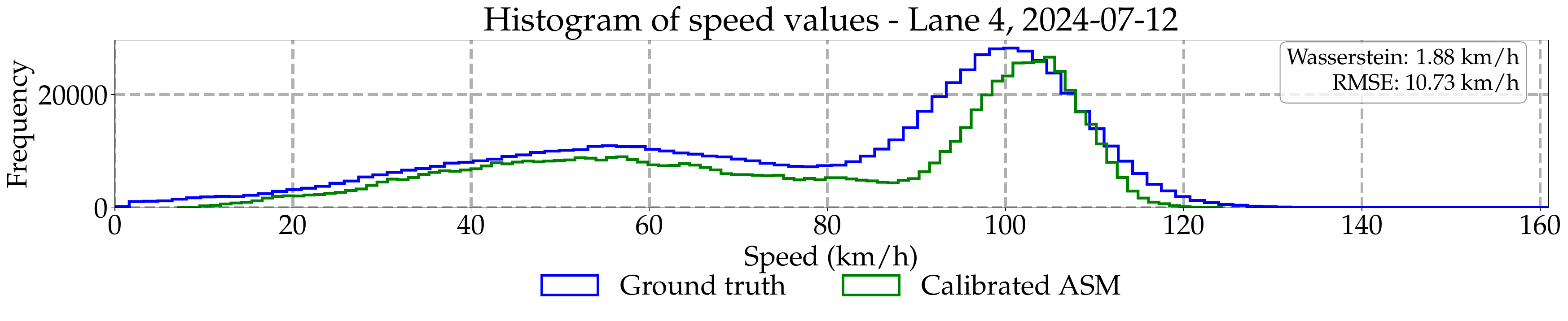}
        \caption{Lane 4 -- July 12}
        \label{fig:l4-12}
    \end{subfigure}
    
    \caption{Comparison of Speed Value Histograms. The left column shows Lane 3 and the right column shows Lane 4, spanning from July 08 to July 12, 2024. Ground truth is shown in blue and calibrated ASM in green.}
    \label{fig:speed_histograms_lane3_4}
\end{figure}

\section*{Acknowledgment}
We appreciate Dr. Xinyu Chen from MIT for his constructive suggestion on improving computational efficiency, and to Austin Coursey and Alex Richardson from Vanderbilt University for their extensive discussions on freeway spatio-temporal data modeling. We also appreciate the GitHub user \texttt{seungwooham} for the feedback on our first implementation at \url{https://github.com/I24-MOTION/VT_tools}. This work
was supported by the National Science Foundation (NSF) under Grant No. 2135579, 2434400 (Work) and the Tennessee Department of Transportation (TDOT) under Grant No. RES2023-20 and Grant No. OTH2023-01F-01. The content is solely the responsibility of the authors and does not necessarily represent the official views of the NSF or TDOT.
\bibliographystyle{ieeetr}
\bibliography{library}

\begin{thebibliography}{10}

\bibitem{treiber2011reconstructing}
M.~Treiber, A.~Kesting, and R.~E. Wilson, ``Reconstructing the traffic state by
  fusion of heterogeneous data,'' {\em Computer-Aided Civil and Infrastructure
  Engineering}, vol.~26, no.~6, pp.~408--419, 2011.

\bibitem{helbing1998jams}
D.~Helbing and M.~Treiber, ``Jams, waves, and clusters,'' {\em Science},
  vol.~282, no.~5396, pp.~2001--2003, 1998.

\bibitem{treiber2002reconstructing}
M.~Treiber and D.~Helbing, ``Reconstructing the spatio-temporal traffic
  dynamics from stationary detector data,'' {\em Cooper@ tive Tr@ nsport@ tion
  Dyn@ mics}, vol.~1, no.~3, pp.~3--1, 2002.

\bibitem{treiber2003adaptive}
M.~Treiber and D.~Helbing, ``An adaptive smoothing method for traffic state
  identification from incomplete information,'' in {\em Interface and transport
  dynamics: Computational Modelling}, pp.~343--360, 2003.

\bibitem{kesting2010datengestutzte}
A.~Kesting and M.~Treiber, ``Datengest{\"u}tzte analyse der stauentstehung
  und-ausbreitung auf autobahnen,'' {\em Stra{\ss}enverkehrstechnik}, vol.~1,
  pp.~5--11, 2010.

\bibitem{treiber2000congested}
M.~Treiber, A.~Hennecke, and D.~Helbing, ``Congested traffic states in
  empirical observations and microscopic simulations,'' {\em Physical review
  E}, vol.~62, no.~2, p.~1805, 2000.

\bibitem{schreiter2010two}
T.~Schreiter, H.~van Lint, M.~Treiber, and S.~Hoogendoorn, ``Two fast
  implementations of the adaptive smoothing method used in highway traffic
  state estimation,'' in {\em 13th International IEEE Conference on Intelligent
  Transportation Systems}, pp.~1202--1208, IEEE, 2010.

\bibitem{schreiter2013vehicle}
T.~Schreiter, ``Vehicle-class specific control of freeway traffic,'' 2013.

\bibitem{mohammadian2021performance}
S.~Mohammadian, Z.~Zheng, M.~M. Haque, and A.~Bhaskar, ``Performance of
  continuum models for realworld traffic flows: Comprehensive benchmarking,''
  {\em Transportation Research Part B: Methodological}, vol.~147, pp.~132--167,
  2021.

\bibitem{he2024efficient}
Y.~He, C.~An, Y.~Jia, J.~Liu, Z.~Lu, and J.~Xia, ``Efficient and robust freeway
  traffic speed estimation under oblique grid using vehicle trajectory data,''
  {\em IEEE Transactions on Intelligent Transportation Systems}, 2024.

\bibitem{wu2024traffic}
F.~Wu, Z.~Cheng, H.~Chen, Z.~Qiu, and L.~Sun, ``Traffic state estimation from
  vehicle trajectories with anisotropic gaussian processes,'' {\em
  Transportation Research Part C: Emerging Technologies}, vol.~163, p.~104646,
  2024.

\bibitem{he2023refining}
Z.~He, ``Refining time-space traffic diagrams: A simple multiple linear
  regression model,'' {\em IEEE Transactions on Intelligent Transportation
  Systems}, vol.~25, no.~2, pp.~1465--1475, 2023.

\bibitem{ji2025stop}
J.~Ji, A.~Richardson, D.~Gloudemans, G.~Zach{\'a}r, M.~Nice, W.~Barbour,
  J.~Sprinkle, B.~Piccoli, and D.~B. Work, ``Stop-and-go wave super-resolution
  reconstruction via iterative refinement,'' {\em Transportation Research Part
  C: Emerging Technologies}, vol.~180, p.~105313, 2025.

\bibitem{benkraouda2020traffic}
O.~Benkraouda, B.~T. Thodi, H.~Yeo, M.~Menendez, and S.~E. Jabari, ``Traffic
  data imputation using deep convolutional neural networks,'' {\em IEEE
  Access}, vol.~8, pp.~104740--104752, 2020.

\bibitem{yang2024advanced}
C.~Yang, {\em Advanced Algorithms for Traffic Data Imputation and City-Wide
  Traffic Dynamics Analysis}.
\newblock PhD thesis, New York University Tandon School of Engineering, 2024.

\bibitem{coursey2024ft}
A.~Coursey, J.~Ji, M.~Quinones~Grueiro, W.~Barbour, Y.~Zhang, T.~Derr,
  G.~Biswas, and D.~Work, ``{FT-AED}: Benchmark dataset for early freeway
  traffic anomalous event detection,'' {\em Advances in Neural Information
  Processing Systems}, vol.~37, pp.~15526--15549, 2024.

\bibitem{li2021multistep}
G.~Li, V.~L. Knoop, and H.~Van~Lint, ``Multistep traffic forecasting by dynamic
  graph convolution: Interpretations of real-time spatial correlations,'' {\em
  Transportation Research Part C: Emerging Technologies}, vol.~128, p.~103185,
  2021.

\bibitem{chen2024macro}
X.~Chen, G.~Qin, T.~Seo, J.~Yin, Y.~Tian, and J.~Sun, ``A macro-micro approach
  to reconstructing vehicle trajectories on multi-lane freeways with lane
  changing,'' {\em Transportation research part C: emerging technologies},
  vol.~160, p.~104534, 2024.

\bibitem{wang2011estimating}
M.~Wang, W.~Daamen, S.~Hoogendoorn, and B.~Van~Arem, ``Estimating acceleration,
  fuel consumption, and emissions from macroscopic traffic flow data,'' {\em
  Transportation research record}, vol.~2260, no.~1, pp.~123--132, 2011.

\bibitem{tsanakas2022generating}
N.~Tsanakas, J.~Ekstr{\"o}m, and J.~Olstam, ``Generating virtual vehicle
  trajectories for the estimation of emissions and fuel consumption,'' {\em
  Transportation Research Part C: Emerging Technologies}, vol.~138, p.~103615,
  2022.

\bibitem{van2010robust}
J.~Van~Lint and S.~P. Hoogendoorn, ``A robust and efficient method for fusing
  heterogeneous data from traffic sensors on freeways,'' {\em Computer-Aided
  Civil and Infrastructure Engineering}, vol.~25, no.~8, pp.~596--612, 2010.

\bibitem{ngsim}
{U.S. Department of Transportation, Federal Highway Administration}, ``Next
  generation simulation (ngsim) vehicle trajectories and supporting data.''
  \url{https://ops.fhwa.dot.gov/trafficanalysistools/ngsim.htm}, 2007.
\newblock Accessed: 2025-04-29.

\bibitem{seo2020evaluation}
T.~Seo, Y.~Tago, N.~Shinkai, M.~Nakanishi, J.~Tanabe, D.~Ushirogochi,
  S.~Kanamori, A.~Abe, T.~Kodama, S.~Yoshimura, {\em et~al.}, ``Evaluation of
  large-scale complete vehicle trajectories dataset on two kilometers highway
  segment for one hour duration: Zen traffic data,'' in {\em 2020 International
  Symposium on Transportation Data and Modelling}, 2020.

\bibitem{chen2018adaptive}
X.~Chen, S.~Zhang, L.~Li, and L.~Li, ``Adaptive rolling smoothing with
  heterogeneous data for traffic state estimation and prediction,'' {\em IEEE
  transactions on intelligent transportation systems}, vol.~20, no.~4,
  pp.~1247--1258, 2018.

\bibitem{chen2024forecasting}
X.~Chen, X.-L. Zhao, and C.~Cheng, ``Forecasting urban traffic states with
  sparse data using hankel temporal matrix factorization,'' {\em INFORMS
  Journal on Computing}, 2024.

\bibitem{coifman2023lwr}
B.~Coifman, B.~Ponnu, and P.~El~Asmar, ``Lwr and shockwave analysis-failures
  under a concave fundamental diagram and unexpected induced disturbances,''
  {\em Transportation Research Part A: Policy and Practice}, vol.~175,
  p.~103766, 2023.

\bibitem{coifman2024microscopic}
B.~Coifman, ``Microscopic discontinuities disrupting hydrodynamic and continuum
  traffic flow models,'' {\em Transportation Research Part B: Methodological},
  vol.~189, p.~103068, 2024.

\bibitem{treiber2013traffic}
M.~Treiber and A.~Kesting, {\em Traffic Flow Dynamics: Data, Models and
  Simulation}.
\newblock Berlin, Heidelberg: Springer, 2013.

\bibitem{ji2024scalable}
J.~Ji, D.~Gloudemans, Y.~Wang, G.~Zach{\'a}r, W.~Barbour, J.~Sprinkle,
  B.~Piccoli, and D.~B. Work, ``Scalable analysis of stop-and-go waves:
  Representation, measurements and insights,'' {\em Transportation Research
  Part C: Emerging Technologies}, vol.~182, p.~105385, 2026.

\bibitem{gloudemans202324}
D.~Gloudemans, Y.~Wang, J.~Ji, G.~Zachár, W.~Barbour, E.~Hall, M.~Cebelak,
  L.~Smith, and D.~B. Work, ``I-24 {MOTION}: An instrument for freeway traffic
  science,'' {\em Transportation Research Part C: Emerging Technologies},
  vol.~155, p.~104311, 2023.

\bibitem{schicktanz2025dlr}
C.~Schicktanz, L.~Klitzke, K.~Gimm, R.~L{\"u}dtke, K.~Liesner, H.~H. Mosebach,
  F.~M. Heuer, A.~Wodtke, and L.~Asbach, ``The dlr highway traffic dataset
  (dlr-ht): Longest road user trajectories on a german highway,'' {\em
  TechRxiv}, 2025.

\bibitem{paszke2019pytorch}
A.~Paszke, ``Pytorch: An imperative style, high-performance deep learning
  library,'' {\em arXiv preprint arXiv:1912.01703}, 2019.

\bibitem{kingma2014adam}
D.~P. Kingma and J.~Ba, ``Adam: A method for stochastic optimization,'' {\em
  arXiv preprint arXiv:1412.6980}, 2014.

\bibitem{thop_pypi}
L.~Zhu, ``thop: A tool to count the flops of pytorch model.''
  \url{https://pypi.org/project/thop/}, 2022.
\newblock Version 0.1.1.post2209072238. Accessed 2026-01-31.

\bibitem{wavetronix_smartsensor_hd}
{Wavetronix}, ``{SmartSensor HD}.''
  \url{https://www.wavetronix.com/products/smartsensor-hd}, 2025.
\newblock Accessed: 2025-05-01.

\bibitem{ji2024virtual}
J.~Ji, Y.~Wang, D.~Gloudemans, G.~Zach{\'a}r, W.~Barbour, and D.~B. Work,
  ``Virtual trajectories for i--24 motion: Data and tools,'' in {\em 2024 Forum
  for Innovative Sustainable Transportation Systems (FISTS)}, pp.~1--6, IEEE,
  2024.

\bibitem{edie1963discussion}
L.~C. Edie {\em et~al.}, ``Discussion of traffic stream measurements and
  definitions.'' Port of New York Authority New York, 1963.

\bibitem{knoop2012quantifying}
V.~L. Knoop, S.~Hoogendoorn, Y.~Shiomi, and C.~Buisson, ``Quantifying the
  number of lane changes in traffic: Empirical analysis,'' {\em Transportation
  research record}, vol.~2278, no.~1, pp.~31--41, 2012.

\bibitem{villani2008optimal}
C.~Villani, {\em Optimal Transport: Old and New}, vol.~338.
\newblock Grundlehren der mathematischen Wissenschaften: Springer, 2008.

\bibitem{ahn2022reproducibility}
K.~Ahn, P.~Jain, Z.~Ji, S.~Kale, P.~Netrapalli, and G.~I. Shamir,
  ``Reproducibility in optimization: Theoretical framework and limits,'' {\em
  Advances in Neural Information Processing Systems}, vol.~35,
  pp.~18022--18033, 2022.

\bibitem{zhang2025real}
Y.~Zhang, Z.~Zhang, J.~Ji, M.~Quinones-Grueiro, W.~Barbour, D.~Gloudemans,
  C.~Weston, G.~Biswas, D.~B. Work, {\em et~al.}, ``Real-world deployment and
  assessment of a multi-agent reinforcement learning-based variable speed limit
  control system,'' {\em arXiv preprint arXiv:2503.01017}, 2025.

\bibitem{yang2022generalized}
C.~Yang, B.~T. Thodi, and S.~E. Jabari, ``Generalized adaptive smoothing using
  matrix completion for traffic state estimation,'' in {\em 2022 IEEE 25th
  International Conference on Intelligent Transportation Systems (ITSC)},
  pp.~787--792, IEEE, 2022.

\end{thebibliography}
}
\end{spacing}
\end{document}